\documentclass[11pt]{article}
\usepackage{amsmath}
\usepackage{bm}
\usepackage{natbib}
\usepackage[colorlinks,citecolor=blue,linkcolor=blue,urlcolor=blue]{hyperref}
\usepackage{latexsym, epsfig, amssymb, amsmath, amsthm,graphicx, mathrsfs, booktabs}
\usepackage{rotating, tabularx, setspace,paralist}
\usepackage{color}
\usepackage[OT1]{fontenc}
\usepackage{lscape}
\usepackage{multirow,multicol}
\usepackage{anysize}

\marginsize{1.1in}{1.1in}{0.55in}{0.8in} %

\makeatletter
\usepackage{accents}
\usepackage{bbm} 
\usepackage[linesnumbered,ruled,vlined]{algorithm2e} 
\usepackage{algorithmic}
\usepackage{subfig}
\usepackage{mathtools}

\usepackage{verbatim} 
\usepackage[utf8]{inputenc}
\usepackage[english]{babel}
\usepackage[T1]{fontenc}
\usepackage[utf8]{inputenc}
\usepackage{authblk}
\usepackage{tikz}
\usepackage{pifont}

\newtheorem{defn}{Definition} 
\newtheorem*{defn*}{Definition} 

\newtheorem{condi}{Condition} 
\newtheorem*{condi*}{Condition} 

\newtheorem{theorem}{Theorem}

\newtheorem{lemma}{Lemma}
\newtheorem*{lemma*}{Lemma}
\newtheorem{proposition}{Proposition}

\theoremstyle{plain}

\newtheorem*{claim*}{Claim}
\newtheorem*{rwr*}{Removed When Ready}

\usepackage{enumitem}
\newlist{steps}{enumerate}{1}
\setlist[steps, 1]{label = Step \arabic*:}
\usepackage{mathtools}

\SetKwInput{KwInput}{Input}                
\SetKwInput{KwOutput}{Output}              

\renewcommand\qquad{\quad}
\renewcommand\dots{\cdots}
\renewcommand\top{T}

\begin{document}
	
\title{FACT: High-Dimensional Random Forests Inference%
\thanks{
Chien-Ming Chi is Assistant Research Fellow, Institute of Statistical Science, Academia Sinica (E-mail: \textit{xbbchi@stats.sinica.edu.tw}). %
Yingying Fan is Centennial Chair in Business Administration and Professor, Data Sciences and Operations Department, Marshall School of Business, University of Southern California, Los Angeles, CA 90089 (E-mail: \textit{fanyingy@marshall.usc.edu}). %
Jinchi Lv is Kenneth King Stonier Chair in Business Administration and Professor, Data Sciences and Operations Department, Marshall School of Business, University of Southern California, Los Angeles, CA 90089 (E-mail: \textit{jinchilv@marshall.usc.edu}). %
This work was supported by NSF Grants DMS-1953356, EF-2125142, DMS-2310981, and  DMS-2324490, and by Grant 111-2118-M-001-012-MY2 from the National Science and Technology Council, Taiwan.
}
\date{November 12, 2023}
\author{Chien-Ming Chi$^1$, Yingying Fan$^2$ and Jinchi Lv$^2$
\medskip\\
Academia Sinica$^1$ and University of Southern California$^2$
\\
} %
}

\maketitle
	
\begin{abstract}
Quantifying the usefulness of individual features in random forests learning can greatly enhance its interpretability. Existing studies have shown that some popularly used feature importance measures for random forests suffer from the bias issue. In addition, there lack comprehensive size and power analyses for most of these existing methods. In this paper, we approach the problem via hypothesis testing, and suggest a framework of the self-normalized feature-residual correlation test (FACT) for evaluating the significance of a given feature in the random forests model with bias-resistance property, where our null hypothesis concerns whether the feature is conditionally independent of the response given all other features. Such an endeavor on random forests inference is empowered by some recent developments on high-dimensional random forests consistency. Under a fairly general high-dimensional nonparametric model setting with dependent features, we formally establish that FACT can provide theoretically justified feature importance test with controlled type I error and enjoy appealing power property. The theoretical results and finite-sample advantages of the newly suggested method are illustrated with several simulation examples and an economic forecasting application.
\end{abstract}
	
\textit{Running title}: FACT
	

\section{Introduction} \label{Sec.new1}

Reliable statistical inference depends on an accurate modeling for the observed data. In modern applications, the data collected is often high-dimensional and exhibits complex relationships between the response and its explanatory features. Such an involved modelling task can be done by the state-of-the-art machine learning methods. One method of particular interest is 
	the random forests~\citep{Breiman2001, biau2016random}, which is an ensemble predictive model making predictions by aggregating the individual predictions from a collection of randomized decision trees. Over the past two decades, a vast amount of research has demonstrated that random forests is reliable in applications from diverse disciplines such as economics~\citep{medeiros2021forecasting, Athey2019, wager2018estimation}, finance~\citep{khaidem2016predicting}, bioinformatics~\citep{qi2012random, diaz2006gene}, biostatistics~\citep{ishwaran2008random}, and multi-source remote sensing~\citep{gislason2006random}. Theoretical endeavors have also proven the consistency of random forests~\citep{Biau2015, biau2012analysis, biau2008consistency} under various nonparametric settings, even in the presence of correlated and high-dimensional features \citep{chi2020asymptotic, klusowski2021universal, syrgkanis2020estimation}.

	Despite the appealing estimation/prediction accuracy, the results output by the random forests can be hard to interpret because of its black-box nature. Much effort has been made to enhance its interpretibaility, mainly using the idea of identifying most significant features in explaining the response. Among the endeavors, permutation inference is a popular approach for accessing variable importance; see, for example,  \cite{altmann2010permutation, hapfelmeier2013new}. The methods therein were designed based on the intuition that the importance of a feature can be assessed by checking the change of some appropriately chosen measure before and after randomly permuting some variables, either the feature to be evaluated~\citep{hapfelmeier2013new} or the response~\citep{altmann2010permutation}, with the former work focusing on testing the marginal importance of features and the latter one considering both the marginal and conditional importance of features. Both methods output p-values for testing the null hypothesis that a pre-chosen feature is unimportant based on some heuristic arguments without formal theoretical guarantees. Another hypothesis test based approach is the prediction difference test proposed in~\cite{mentch2016quantifying}, which assesses the significance of an explanatory feature based on the prediction difference between the full model and the reduced model without using the feature to be tested. There  exist only limited theoretical justifications on the proposed tests along this line of work.

The relative feature importance evaluation is a more general framework than the testing approach reviewed above for measuring the feature importance, where the goal is to find out the relative importance of a feature to other features in predicting the response. Yet, relative importance measures are less informative because they do not quantitatively assess the significance of explanatory features as p-values do. Some popularly used random forests feature importance measures include the mean decrease accuracy (MDA)~\citep{Breiman2001}, the mean decrease in impurity (MDI)~\citep{Breiman2002}, and the conditional permutation importance (CPI)~\citep{strobl2008conditional}.  The MDA and CPI are based on the permutation approach, whereas the MDI evaluates a feature by the contributions of all of its associated tree branch splits toward explaining the variation of the response. These intuitive approaches are popularly used for evaluating the relative feature importance for random forests. To name a few recent works along this line, see, e.g., \cite{li2019debiased, loecher2020unbiased, mao2022nonparametric, zhou2021unbiased, benard2021mda}. Although the MDI and MDA may be intuitive, they can suffer from the bias issue toward spurious features when feature dependence is present, as revealed in the literature ~\citep{strobl2008conditional, genuer2010variable, nicodemus2009predictor}. Various modifications have been proposed to correct such bias. The CPI and a few other works \citep{loecher2020unbiased, zhou2021unbiased, benard2021mda, agarwal2023mdi} alleviate this bias issue of the MDA and MDI. However, in general nonparametric model settings and with high dimensionality, the bias-resistance properties of these measures have not been fully analyzed from a theoretical standpoint.

To better motivate the need for a more bias-resistant random forests feature importance measure, we have conducted a simulation study comparing MDI, MDA, and CPI with our newly proposed measure, which we abbreviate as FACT for the ease of presentation.
We provide in Table~\ref{tab:5}  the simulation results for examining the spurious effects on MDI, MDA, CPI, and FACT. Specifically, we calculate the percentage of times that each of these measures ranks the null feature $X_{2}$ higher than the relevant feature $X_{11}$, where $X_{2}$ is correlated with another relevant feature $X_{1}$ in model \eqref{Y1}. A larger entry in Table~\ref{tab:5} suggests a stronger spurious effect and thus means worse performance of the corresponding measure.

		\begin{table}
		\begin{center}
		\begin{tabular}[t]{cc} 
			\begin{tabular}[t]{ p{1.4em}   | c|c| c|c}
    \hline
				&
    {MDI$_{2}>$ MDI$_{11}$} & 
    {MDA$_{2}>$ MDA$_{11}$}&
    {CPI$_{2}>$ CPI$_{11}$} & 
    {FACT$_{2}>$ FACT$_{11}$}\\
				\hline					
				I &0.08 & 0.03  & 0.03 & 0.01 \\
				\hline
				II&0.99 & 0.99 & 0.96 & 0.26  \\
				\hline
				III &1.00 & 1.00 & 0.29 &0.00  \\
				\hline				
			\end{tabular}	
			\ \ 
		\end{tabular}
		\captionof{table}{Spurious effects of random forests feature importance measures and the FACT statistic under model~\eqref{Y1} from Section \ref{Sec7.3}: I) $(n, p, \lambda) = (60, 30, 0.1)$, II) $(n, p, \lambda) = (60, 30, 0.8)$, and III) $(n, p, \lambda) = (200, 30, 0.8)$.  Each entry represents the fraction of times out of 100 simulation repetitions when the feature importance measure for the null feature $X_{2}$ is larger than that for relevant feature $X_{11}$. The null feature $X_{2}$ is correlated with the strong linear component $X_{1}$ in model \eqref{Y1}.  Larger entries indicate stronger spurious effects.}\label{tab:5}
	\end{center}
	\end{table}

It is seen that when features have very low dependence (setting I), all four measures behave well in ranking relevant feature $X_{11}$ over the spurious null feature $X_2$. However, when feature dependence is high in cases II and III, the MDI and MDA can be seriously biased toward the spurious feature $X_2$. The well-established bias-resistant importance measure CPI performs much better than MDI and MDA with a larger sample size in case III, but even CPI shows significant bias in case II. In contrast, our FACT statistic greatly alleviates such a bias issue, underscoring its advantage. Additionally, we assessed the FACT's ability to resist bias by comparing it with the recently proposed bias-resistant test statistic, Generalized Covariance Measure (GCM,~\cite{shah2020hardness}), in Tables~\ref{tab:4}--\ref{tab:4.gcm}; the GCM is briefly introduced below. While a comprehensive empirical comparison with various other recently proposed bias-resistant versions of MDI and MDA, as mentioned earlier, would be intriguing, we leave such in-depth discussion for future research.


We next introduce our FACT framework, and start with reviewing the GCM~\citep{shah2020hardness}, which is the building block of FACT. Let $Y\in \mathbb R$ be the response of interest and $\boldsymbol X = (X_1, \cdots, X_p)^T\in \mathbb R^p$  the $p$-dimensional feature vector.  We evaluate the importance of the $j$th feature by formally testing the null hypothesis  
	\begin{equation}
		\label{null3}
		H_{0}: \textnormal{ The } j\textnormal{th feature $X_j$ is a null feature,} 
	\end{equation}
	where our formal definition of null features is given in Definition \ref{null1} in Section \ref{Sec.new4.1} and focuses on the conditional independence of $X_j$ with $Y$ given all remaining features \citep{CandesFanJansonLv2018, strobl2008conditional}. Let independent and identically distributed (i.i.d.) observations of the response and  feature vector $\{(Y_{i}, \boldsymbol{X}_{i})\}_{i=1}^{n}$ be given, and  assume temporarily an errorless model training where the residuals after model training are $R_{1i} \coloneqq g_{1}(Y_{i}) - \mathbb{E}(g_{1}(Y_{i})|\boldsymbol{X}_{-ij})$ and $R_{2i} \coloneqq g_{2}(X_{ij}) - \mathbb{E}(g_{2}(X_{ij})|\boldsymbol{X}_{-ij})$ for all $i$, with $\boldsymbol{X}_{-ij}$ the $i$th observation with the $j$th feature $X_{ij}$ excluded. The GCM statistic is given by
	\begin{equation}
		\label{fact0} \frac{\sum_{i=1}^{n} R_{1i} R_{2i}}{\sqrt{ \sum_{i=1}^{n} (R_{1i} R_{2i} -  n^{-1}\sum_{i}^{n} R_{1i} R_{2i})^2}}.
	\end{equation}
Observe that $R_{1i}R_{2i}$, $i=1,\cdots, n$, are i.i.d. with $\mathbb{E}(R_{11}R_{21}) = 0$ under the null hypothesis. Hence, with regularity conditions assumed, the  above GCM test \eqref{fact0} is asymptotically normal. 
Directly applying the above GCM test for evaluating random forests feature importance can yield inflated type I error and no power (cf. Table \ref{tab:4.gcm} in our simulation section). The unsatifactory performance is mainly caused by the slow convergence rate of random forests in high-dimensional setting \citep{chi2020asymptotic} and the possible nonlinear dependence of response on features. Our FACT framework is designed specifically for random forests to address the challenges in GCM and other popularly used existing measures.

FACT constructs the conditional means $\mathbb E(g_1(Y_i)|\boldsymbol{X}_{-ij})$ and $\mathbb E(g_2(X_{ij})|\boldsymbol{X}_{-ij})$ by using the random forests models learned from an independent training data. Then the GCM test \eqref{fact0} is calculated using an inference sample with the learned random forests models. The use of independent training sample avoids the overfitting issue. To alleviate the bias caused by the slow random forests convergence, we use imbalanced samples where the random forests training sample size is much larger than that of the inference sample. In the absence of independent training sample, we adopt the idea of sample splitting and cross-fitting for effective use of the data, where the entire data is  split into $K\geq 3$ equal-sized folds with one fold used for inference and the rest used for training. This yields $K$ sets of inference result with each corresponding to one fold of the data. To aggregate these inference results, we adopt the idea of e-value statistic~\citep{wang2022false} and take advantage of the property that the average of e-values still yields a valid e-value. A rejection rule based on the aggregated e-value is then constructed. In practical implementation, we suggest the out-of-bag (OOB) random forests estimation for faster implementation. Our numerical studies in this paper are all conducted using the OOB implementation. We prove theoretically via a nonasymptotic analysis that FACT based on independent and unbalanced training sample yields a valid size controlled below the target level. We also use a simulation study to demonstrate that the OOB implementation achieves the type I error control in finite samples. 

To enhance the power of FACT, we employ different  transformations $g_{1l}(\cdot)$ and $g_{2l}(\cdot)$, $l=1,\cdots, L$, to both the response and features. The e-values obtained from different transformations are averaged to obtain the final e-value for testing \eqref{null3}. We showcase the nonasymptotic power analysis by considering some specific transformation functions. Our results clearly reveal the advantage of using multiple pairs of transformations in the sense of capturing a richer family of dependence structures between the response and features, especially when compared to FACT without any data transformation (i.e., the identity transformation is used). We justify these theoretical findings and demonstrate the finite-sample power results of FACT through simulation studies.

We demonstrate the practical applicability of FACT test by applying it to analyze the macroeconomic data FRED-MD~\citep{mccracken2016fred}, where the goal is to assess the importance of a set of macroeconomic variables in predicting the inflation rate. To address the common concern of nonstationarity of time series over a long time period, we break the entire time series into small rolling windows, each with only two years data of $24$ monthly  observations. Motivated from this application, we also propose the stable FACT that aims to address the reproducibility issue related to small sample size and intrinsic randomness in random forests and sample splitting. We demonstrate the performance of stable FACT by using both simulation study and applying it to the FRED-MD data.


The rest of the paper is organized as follows.
Section \ref{Sec2} introduces the model setting and provides a brief overview of high-dimensional random forests consistency. We introduce the framework of the FACT test for high-dimensional random forests inference of feature importance in Section \ref{Sec.new3}. Section \ref{Sec.new4} presents the nonasymptotic theory of the FACT test from both perspectives of the size and power.  Sections \ref{Sec7} and \ref{Sec8} present several
simulation and real data examples illustrating the finite-sample performance and utility of
our newly suggested method. All the proofs and technical details are provided in the Supplementary Material.

\section{High-dimensional random forests} \label{Sec2}

In this section, we will introduce some necessary technical background on high-dimensional random forests consistency, which will empower the nonasymptotic theory for our framework of the FACT test to be presented in Section \ref{Sec.new4}. Denote by $(\Omega, \mathcal{F}, \mathbb{P})$ the underlying probability space, $Y$ a scalar response, and $\boldsymbol{X} \coloneqq (X_{1}, \dots, X_{p})^{\top}$ a $p$-dimensional random feature vector taking values in $\mathbb{R}^p$. Assume that we are given an inference sample of independent and identically distributed (i.i.d.) observations $\{(\boldsymbol{X}_{i}, Y_{i})\}_{i=1}^{n}$, where $\boldsymbol{X}_{i}\coloneqq(X_{i1},\dots, X_{ip})^{\top}$, $(\boldsymbol{X}_{i}, Y_{i})$ and $(\boldsymbol{X}, Y)$ have the same distribution, and $(\boldsymbol{X}, Y)$ is independent of all the observations. Further, assume that we have an independent training sample $\mathcal{X}_{0} = \{\boldsymbol{U}_{i}, V_{i}\}_{i=1}^{N}$ from the same distribution as $(\boldsymbol{X}, Y)$, where $\boldsymbol{U}_{i} = (U_{i1}, \cdots, U_{ip})^T$.

We first consider a single null hypothesis \eqref{null3} where $1\le j\le p$ is a pre-chosen covariate index. The problem of multiple hypotheses testing will be addressed in Section \ref{Sec4.3}. As mentioned in the Introduction, we will construct random forests estimates of $\mathbb{E}(g_{1}(Y)  \vert  \boldsymbol{X}_{-j})$ and $\mathbb{E}(g_{2}(X_{j})  \vert  \boldsymbol{X}_{-j})$ for testing null hypotheses \eqref{null3}, where $\boldsymbol{X}_{-j} \coloneqq (X_{1}, \dots, X_{j-1}, X_{j+1}, \dots, X_{p})^{\top}$. As shown in Section~\ref{Sec.new3}, the inference sample of size $n$ will be used for calculating the FACT statistics, while the training sample $\mathcal{X}_{0}$ of size $N$ will be employed for constructing the underlying random forests estimates. Denote by $\widehat{Y}(\boldsymbol{X}_{-j})$ and $\widehat{X}(\boldsymbol{X}_{-j})$ the random forests estimates of $\mathbb{E}(g_{1}(Y)  \vert  \boldsymbol{X}_{-j})$ and $\mathbb{E}(g_{2}(X_{j})  \vert  \boldsymbol{X}_{-j})$, respectively,  constructed using the independent training sample $\{\boldsymbol{U}_{-ij}, V_{i}\}_{i=1}^{N}$, where $\boldsymbol{U}_{-ij}\coloneqq (U_{i1}, \dots, U_{i, j-1}$, $U_{i, j+1}, \dots, U_{ip})^{\top}$. To ensure valid statistical inference using the random forests estimates, we impose the regularity conditions below on the random forests consistency in the high-dimensional nonparametric regression setting.

\begin{condi}\label{A1}
Assume that $\mathbb{E}\{ \mathbb{E}(g_{1}(Y) | \boldsymbol{X}_{-j})  - \widehat{Y}(\boldsymbol{X}_{-j})\}^{2} \le B_{1}$ for some small $B_{1}>0$.
\end{condi}
	
    \begin{condi}\label{A7}
    Assume that $\mathbb{E}\{\mathbb{E}(g_{2}(X_{j}) | \boldsymbol{X}_{-j}) - \widehat{X}( \boldsymbol{X}_{-    j})\}^{2} \le B_{2}$ for some small $B_2 > 0$.
    \end{condi}


Conditions \ref{A1} and \ref{A7} above are similar in nature so we only provide discussions on Condition \ref{A1}. There is a growing recent literature on the random forests consistency, which amounts to Condition~\ref{A1} with $g_1(y)=y$, $X_{j}$ being a null feature (see Definition \ref{null1} in Section~\ref{Sec.new4.1}), and the consistency rate $B_{1} = o(1)$ depending on the training sample size $N$. For example, \cite{Biau2015,chi2020asymptotic, syrgkanis2020estimation, klusowski2021universal} established the $\mathbb{L}^{2}$-consistency of random forests with decision trees grown by the original Breiman's classification and regression tree (CART) splitting criterion \citep{Breiman2001, Breiman2002} under various settings of the nonparametric regression model $Y= m(\boldsymbol{X}) + \varepsilon$. 
Here, $m(\cdot)$ represents the underlying true regression function, and $\varepsilon$ is the model error that is independent of feature vector $\boldsymbol{X}$ and has mean zero and finite variance. In particular, by assuming that the true regression function $m(\cdot)$ and the distribution of feature vector $\boldsymbol{X}$ satisfy a condition called the sufficient impurity decrease (SID), \cite{chi2020asymptotic} established the high-dimensional random forests consistency rates in a general nonparametric model setting with dependent features; their results ensure that Condition~\ref{A1} holds with $g_1(y)=y$, consistency rate $B_{1}= O(N^{-c})$, and feature dimensionality $p=O(N^{K_{0}})$ for some constants $c, K_{0}>0$. 
In addition to the aforementioned works, the consistency of many variants of random forests has also been investigated in the recent literature. These variants usually consider models of decision trees that are grown by certain splitting protocols other than the original CART criterion; see, e.g., \cite{Biau2015, chi2020asymptotic, biau2016random, klusowski2021universal} for detailed overviews.

\section{FACT for high-dimensional random forests inference} \label{Sec.new3}

 \subsection{The FACT algorithm}\label{fact-algorithm}
We now introduce the main ideas for the FACT framework. Our framework is built upon the GCM statistic and unitizes various techniques for debiasing and  power enhancement. We first introduce Algorithm~\ref{Algorithm1} and subsequently discuss how to enhance the selection power when dealing with nonlinear features in Algorithm~\ref{Algorithm2}.

\bigskip
    \begin{algorithm}[H]\label{Algorithm1}
		\SetAlgoLined
\KwIn{%
         Transformation functions $g_{1}:\mathbb{R}\longmapsto\mathbb{R}$ and $g_{2}:\mathbb{R}\longmapsto\mathbb{R}$, inference sample $\{Y_{i}, \boldsymbol{X}_{i}\}_{i=1}^n$, and training sample $\{V_{i}, \boldsymbol{U}_{i}\}_{i=1}^{N}$ for random forests model fitting. A covariate of interest $X_{j}$ and a  tuning parameter $0\le \epsilon \le 1$.
         }%
\KwOut{%
An e-value statistic denoted as $e_{j}(g_{1}, g_{2})$.
}%

The response vector $(Y_1,\cdots, Y_n)^T$ and each  covariate vector $(X_{1j},\cdots, X_{nj})^T$, $j=1,\cdots,p$, are centered and standardized to have mean zero and unit variance.

Let $H_{1}, \dots, H_{K}$ be a partition of the index set $\{1, \dots, n\}$ satisfying that  $H_{k}\cap H_{l} = \emptyset$ and $|\# H_{k} - \# H_{l}|\le 1$ for all $\{k, l\} \subset \{1, \dots,K\}$ with $k\not= l$.

For each $k \in \{1,\dots ,K\}$, define the FACT test statistic as
\begin{equation}\label{E5}
\textnormal{F}_{j}^{(k)} = \widehat{\sigma}_{j}^{-1}\sum_{i\in H_{k}}d_{i}  \ \text{ with } \widehat{\sigma}_{j}^{2} = \sum_{i\in H_{k}} (d_{i} - (\# H_{k})^{-1}\sum_{i\in H_{k}}d_{i} )^2 ,
\end{equation}
where $d_{i} =  [g_{1}(Y_{i}) - \widehat{Y}(\boldsymbol{X}_{-ij} )]  [g_{2}(X_{ij}) - \widehat{X}(\boldsymbol{X}_{-ij} ) )]$. The estimate $\widehat{Y}: \mathbb{R}^{p-1} \longmapsto\mathbb{R}$ is obtained by regressing $\{g_{1}(V_{i})\}_{i=1}^{n}$ on $\{\boldsymbol{U}_{-ij}\}_{i=1}^{n}$, and $\widehat{X}: \mathbb{R}^{p-1} \longmapsto\mathbb{R}$ is obtained by regressing $\{g_{2}(U_{ij})\}_{i=1}^{n}$ on $\{\boldsymbol{U}_{-ij}\}_{i=1}^{n}$, both using random forests regression. 

The p-value and e-value of the FACT test statistic $\textnormal{F}_{j}^{(k)}$, and the average of e-values are defined as\footnote{The statistic F$_{j}^{(1)}$ defined in \eqref{E5} with $K=1$ and transformation functions $g_{1}(x) = g_{2}(x) = x$ is the GCM statistic given in \cite{shah2020hardness}.}
 \begin{equation}
     \label{E5.3}
     P_{j}^{(k)} \coloneqq 2\Phi(-|\textnormal{F}_{j}^{(k)}| ), \qquad e_{j}^{(k)} \coloneqq (P_{j}^{(k)} \vee \epsilon)^{-\frac{1}{2}} - 1, \qquad e_{j}(g_{1}, g_{2}) \coloneqq K^{-1}\sum_{k=1}^K {e}_{j}^{(k)},
 \end{equation}
 respectively, where $\Phi(t)$ is the standard Gaussian cumulative distribution function (CDF).
\caption{{FACT} }
	\end{algorithm}
    \bigskip

Compared to GCM, which is a generic method designed for testing variable conditional independence, our FACT method incorporates some additional techniques specific to random forests to reduce the bias. To gain some insights into the bias issue, let us consider the case of $K=1$ in Algorithm \ref{Algorithm1}. Under the null hypothesis, we can quantify the bias of $\textnormal{F}_j^{(1)}$ defined in \eqref{E5} as
\begin{equation*}
		\begin{split}\label{bias2}	
			\textnormal{Bias}(N) & \coloneqq \mathbb{E}\Big\{n^{-\frac{1}{2}}\sum_{i=1}^{n} \big[g_{1}(Y_{i}) - \widehat{Y}(\boldsymbol{X}_{-ij}) \big] \big[g_{2}(X_{ij}) - \widehat{X}( \boldsymbol{X}_{-ij}) \big] \Big\}\\
			& =  \sqrt{n} \mathbb{E}\big\{ \big[\mathbb{E}(g_{1}(Y)|\boldsymbol{X}_{-j}) - \widehat{Y}(\boldsymbol{X}_{-j}) \big] \big[\mathbb{E}(g_{2}(X_{j}) |\boldsymbol{X}_{-j}) - \widehat{X}( \boldsymbol{X}_{-j}) \big]  \big\}
	\end{split}\end{equation*}	 
	up to a bounded factor $(\widehat{\sigma}_{j})^{-1}$ in a probabilistic sense. Here, recall that $\mathcal{X}_{0}$ denotes the training sample of size $N$ for constructing $\widehat{Y}(\cdot)$. Assume the use of independent training sample and that Condition~\ref{A1}--\ref{A7} hold with consistency rates $B_{1}$ and $B_{2}$. Simple calculations show that 
	 \begin{equation*}
	     \label{analysis.rate.1}
	     \textnormal{Bias}(N) \le  \sqrt{B_{1}B_{2}n}.
	 \end{equation*}
As discussed in the last section, the random forests consistency rates $B_1$ and $B_2$ are both of order  $N^{-c}$ with $c\in (0,1)$ some constant. This result indicates that having imbalanced sample sizes with $N> nK^{-1}$ can help control the bias. 
In Section \ref{Sec7}, we will demonstrate the bias issue of the  FACT  statistic without using imbalanced samples by a simulation study.  

In practice, the independent samples $\{Y_i, \boldsymbol{X}_i\}_{i=1}^N$ and $\{V_i, \boldsymbol{U}_i\}_{i=1}^n$ in Algorithm \ref{Algorithm1} can be obtained by splitting the entire sample into two equal-sized subsamples. In such a case, we need to use $K\geq 2$  to ensure that each $F_j^{(k)}$ is constructed based on imbalanced training and inference samples. FACT also uses the e-value averaging method to aggregate information across the $K$ FACT statistics. 
Another distinction of FACT from GCM test is the use of transformations $g_1(x)$ and $g_2(x)$, which is for power enhancement and will be discussed in detail in the next section.

 There are alternative methods other than sample splitting to construct the training and inference samples. We discuss two possibilities here. First, the $K$-fold cross-fitting can be used, where for each $k=1,\cdots, K$, we take $\{1, \dots, n\}\backslash H_{k}$ as the training sample for fitting random forests models, and then construct $F_{j}^{(k)}$ based on $H_{k}$ as in \eqref{E5}. Here, we need to choose $K\geq 3$ to ensure imbalanced training and inference sample sizes. Compared to sample splitting, cross-fitting is less demanding in sample size but has a higher computational cost, noting that $K$ pairs of random forests models need to be trained. 
 
 Second, we can use the out-of-bag (OOB) prediction to construct the random forests predictions $\widehat Y$ and $\widehat X$. 
Assume that there are $N_{\text{oob}}$ trees in the random forests. For each $k = 1, \ldots, N_{\text{oob}}$, denote by $a_{k} \subset \{1, \ldots, n\}$ the random subsamples used for training the $k$th decision tree. 
For each observation indexed by $i = 1, \ldots, n$, let $A(i) \subset \{1, \ldots, N_{\text{oob}}\}$ be the set such that $i \notin a_{k}$ for each $k \in A(i)$. This means that for each $1 \leq i \leq n$, the set $A(i)$ contains all the decision trees grown without using the $i$th observation. Naturally, the OOB prediction for the $i$th observation is the empirical average of predicted values given by all decision trees from set $A(i).$ Thus, for each observation $i=1, \ldots, n$, we can form OOB estimates $\widehat{Y}(\boldsymbol{X}_{-ij})$ and $\widehat{X}(\boldsymbol{X}_{-ij})$ and use them to calculate $F_j^{(k)}$ in \eqref{E5} of Algorithm \ref{Algorithm1}. It is seen that the OOB estimate is intended to disentangle the dependency between the training and inference data points, making OOB an alternative to sample splitting. With sufficiently many trees in the random forests model, each OOB estimate is constructed by averaging over a large number of tree estimates and thus, is expected to provide stable results with just one fitting of the random forests models. 
Compared to sample splitting and $K$-fold cross fitting, the OOB implementation offers a practical compromise between computational cost and estimation efficiency (in terms of sample size). Our applications in Sections~\ref{Sec7}--\ref{Sec8} are implemented through the OOB approach.

\subsection{Selection power enhancement}\label{fact.e.power}

The purpose of using feature transformations in Algorithm \ref{Algorithm1} is to increase the power of FACT in identifying important features, especially when data exhibits nonlinear dependency. Feature transformations are common practice in statistical data analyses. Depending on the characteristics of data, practitioners can choose different forms of the transformation.

For example, for $g_{1}(x)$ and $g_{2}(x)$ in Algorithm~\ref{Algorithm1}, we may consider the indentity transformation, denoted as $g(x) = x$, or its quadratic counterpart, $g(x) = x^2$. Section~\ref{Sec4.2} demonstrates that the quadratic transformation is crucial for achieving nontrivial selection power in identifying important features with quadratic or interactive effects. To take advantage of different transformations, in Algorithm~\ref{Algorithm2} below, we will introduce the FACT statistic that aggregates different transformation functions for improving the selection power.

Besides the identity and quadratic transformations, inspired in part by neural network architectures like the long short-term memory networks~\citep{hochreiter1997long}, we have conducted experiments with the hyperbolic tangent function and its quadratic form, defined as $g(x) = \tanh(x)$ and $g(x) = (\tanh(x))^2$, respectively, where $\tanh(x) = \frac{1 - e^{-2x}}{1 + e^{-2x}}$. See Algorithm~\ref{Algorithm2} for details.
In our simulation experiments detailed in Section~\ref{Sec7.2}, we have found that the application of the hyperbolic tangent transformation consistently improves the power of FACT compared to that with no transformation. Furthermore, the outcomes from our real applications demonstrate that our FACT inference with the hyperbolic tangent transformation is indeed able to identify significant macroeconomic time series variables even with just two years of monthly observations.

We have also investigated other squashing functions, including the sigmoid function. In our initial trials, the hyperbolic tangent transformation consistently yielded stable and reliable results compared to the other options. While conducting in-depth experiments to determine the most suitable transformations could offer further insights, we leave this for future research.

\bigskip
    \begin{algorithm}[H]\label{Algorithm2}
		\SetAlgoLined

The power enhanced e-value statistic is defined as
 $$\textnormal{FACT}_{j} \coloneqq L^{-1}\sum_{l=1}^{L} e_{j}(g_{1l}, g_{2l})  ,$$
where each $e_{j}(g_{1l}, g_{2l})$ is given as in \eqref{E5.3} with user-defined transformation functions $g_{1l}:\mathbb{R}\longmapsto\mathbb{R}$ and $g_{2l}:\mathbb{R}\longmapsto\mathbb{R}$ for $l\in \{1, \dots ,L\}$. An example is that $g_{11}(x) = g_{12}(x) = g_{21}(x) = \tanh(x)$ and $g_{22}(x) = (\tanh(x))^2$ with $L=2$, where $\tanh(x) = \frac{1 - e^{-2x}}{1 + e^{-2x}}$.
  
\bigskip
\caption{{FACT with multiple transformation functions} }
	\end{algorithm}
    \bigskip

\section{Nonasymptotic theory of FACT} \label{Sec.new4}

 To simplify the technical presentation, all theoretical results in this section consider the scenario when an independent training sample  $\mathcal{X}_{0} = \{(\boldsymbol{U}_{i}, V_{i})\}_{i=1}^{N}$ is available, and the response and covariates are unnormalized (i.e., Step 1 in Algorithm \ref{Algorithm1} is skipped).

\subsection{The analysis of FACT under null hypothesis}
	\label{Sec.new4.1}

We investigate the performance of FACT under the null hypothesis \eqref{null3}, where the definition of the null feature is formally given below.
\begin{defn}\label{null1}
The $j$th feature $X_{j}$ is said to be a null feature if  $X_{j}$ is conditionally independent of response $Y$ given all remaining features $\boldsymbol{X}_{-j}$.
\end{defn}

 Theorem~\ref{theorem3} below requires Conditions~\ref{A1}--\ref{A7}, which assume high-dimensional consistency rates $B_{1}$ and $B_{2}$ for the random forests estimates. See Section~\ref{Sec2} for how the consistency rates depend on the training sample size, and how Conditions~\ref{A1}--\ref{A7} places some implicit constraints on the underlying distributions of the feature vector and the growth of feature dimensionality in the nonparametric model setting. 
 We also need Condition~\ref{A3} below on  feature dependency structure as well as some regularity conditions. 
 
	\begin{condi}\label{A3}
		The measurable transformation $g_{2}(\cdot)$ is bounded between $0$ and $1$ on its domain. In addition,  $\textnormal{Var}(g_{2}(X_{j}) | \boldsymbol{X}_{-j})\ge \varsigma_{1}$, $\textnormal{Var}(g_{1}(Y)|\boldsymbol{X})\ge \varsigma_{2}$, and $\textnormal{Var}(g_{1}(Y) | \boldsymbol{X}_{-j})\le D$ almost surely, and  $\mathbb{E} [g_{1}(Y)]^4 \le D_{2}$ for some constants $\varsigma_{1}, \varsigma_{2}, D, D_{2} > 0$.  
	\end{condi}

  Condition~\ref{A3} is used to obtain the universal lower and upper bounds for the population variances of the proposed statistics; see Lemma~\ref{lower.var.bounds} in Section~\ref{Sec.newC.32} of the Supplementary Material for details. The lower and upper bounds for $g_2(\cdot)$ can be replaced with any other constants $M_{1}< M_{2}$, respectively. 

	\begin{theorem}\label{theorem3}
	For all large $n$,  all consistency rates $0<B_{1},B_{2}<1$, and each $1\le j\le p$ such that \textnormal{1)} Conditions~\ref{A1}--\ref{A7} hold for random forests estimates constructed using the independent training sample $\mathcal{X}_{0}$, \textnormal{2)} all transformation function pairs $(g_{1}, g_{2})$ and $(g_{1l}, g_{2l})$, $l=1,\cdots, L$, satisfy Condition \ref{A3}, and \textnormal{3)} $X_{j}$ is a null feature, we have that for some $C>0$ and each $\textnormal{F}_{j}^{(k)}$ with $1\le k\le K$ defined in \eqref{E5},
 \begin{equation*}\label{asmy.1}
		\mathbb{P}(|\textnormal{F}_{j}^{(k)}| > t ) \le 2\Phi(-t) + \frac{16c}{5\sqrt{\varsigma_{2}}\varsigma_{1}}+ C(n^{-1/4} + B_{1}^{1/4}+ B_{2}^{1/4}) + (-\log{(B_{1}B_{2})})^{-1},
	\end{equation*}
 where $c  = tn^{-1/4}\log{n} + (2t + 1)( 2B_{1}^{1/4} + B_{2}^{1/4}) + \sqrt{nB_{1}B_{2}} (-\log{(B_{1}B_{2})})$. Moreover, if $(B_{1}+ B_{2})(\log{n})^2\sqrt{n} = o(1)$ is additionally assumed and the tuning parameter satisfies that $\epsilon = (\log{n})^{-1}$, we have that 
 $$\mathbb{E} (e_{j}(g_{1}, g_{2})) \le 1 \  \text{ and } \  \mathbb{E} (\textnormal{FACT}_{j}) \le 1,$$
 where $e_{j}(g_{1}, g_{2})$ and $\textnormal{FACT}_{j}$ are given in \eqref{E5.3} and Algorithm~\ref{Algorithm2}, respectively.		
	\end{theorem}

Theorem~\ref{theorem3} above ensures that for testing the null hypothesis \eqref{null3}, the rejection rule of 
 \begin{equation}
     \label{null.2}
     \textnormal{FACT}_{j} \ge \alpha^{-1}
 \end{equation} 
can have a valid size of $\alpha \in (0,1)$ because 
 $\mathbb{P}(\textnormal{FACT}_{j} \ge \alpha^{-1}| H_0)\le \alpha$
 by the Markov inequality. Note that the above rejection rule is for testing a single null hypothesis \eqref{null.2}. We will discuss the case of simultaneous hypothesis testing later in Section~\ref{Sec4.3}.

	\subsection{Power analysis} \label{Sec4.2}

For the power analysis of FACT, we start with defining a population quantity that can be used to measure the signal strength of features
	\begin{equation} 
	\begin{split}
	\label{new.eq.sec4.2.001}	
	\kappa_{l} & \coloneqq \mathbb{E}\big\{[g_{1l}(Y) - \mathbb{E}( g_{1l}(Y) \vert  \boldsymbol{X}_{-j})] [g_{2l}(X_{j}) - \mathbb{E}(g_{2l}(X_{j}) | \boldsymbol{X}_{-j}) ]\big\}
	\end{split}\end{equation}
	for each $1\le l \le L$, where $L$ denotes the number of different transformation pairs.  Here, the dependence  of $\kappa_{l}$ on $j$ is dropped to simplify the notation.  Since the size and power are two sides of the same coin, we conduct the power analysis with the same regularity conditions as in the last section. As such, the magnitude of $\kappa_l$, which is the conditional covariance of $g_{1l}(Y)$ and $g_{2l}(X_j)$, is of the same order of the corresponding conditional correlation (which is unit-free) in view of Condition \ref{A3}, because the corresponding variances are bounded.

To illustrate the idea of enhancing selection power by multiple transformations, we consider the specific transformations for our technical analysis in this section
 \begin{equation}
 \begin{split}    
     \label{case.1}
     & g_{11}(x) = x, \qquad g_{12}(x) = x ,\qquad
      \dots,  g_{1L}(x) = x, \\
     & g_{21}(x) = x,\qquad g_{22}(x) = x^2, \qquad \dots, g_{2L}(x)=x^L.
     \end{split}
 \end{equation} 
 We provide Theorem~\ref{theorem7} below for analyzing the selection power given nonzero $\kappa_{l}$'s.
 Similar to Theorem \ref{theorem3}, we consider FACT without the centering and standardization step in  Algorithm \ref{Algorithm1} in Theorem~\ref{theorem7}, where we also fix $L=2$ (which is used in our numerical studies) and the tuning parameter $\epsilon = 0$ in Algorithm \ref{Algorithm2} in calculating the FACT statistic.

	\begin{theorem}\label{theorem7}
	Assume that $0\le X_{l}\le 1$ for each $l\in\{1, \dots, p\}$. For all large $n$ and each $1\le j \le p$ such that \textnormal{1)} Conditions~\ref{A1}--\ref{A7} hold with $B_{1}B_{2}n +B_{1} + B_{2}\le 1$ for random forests estimates based on the independent training sample $\mathcal{X}_{0}$, \textnormal{2)} both transformation function pairs $(g_{11}, g_{21})$ and $(g_{12}, g_{22})$ satisfy Condition \ref{A3}, and \textnormal{3)} $|\kappa_{1}| + |\kappa_{2}|>0$, we have  that  
		\begin{equation*}\label{fact.general.2}
			\begin{split}	
   \log{\big[\mathbb{E}(\textnormal{FACT}_{j})\big]}
                 & =O\left[ \frac{n}{K} \times (|\kappa_{1}|\vee |\kappa_{2}|)^2\right],                 
			\end{split}
		\end{equation*}
		where constant integer $K\ge 1$ and $\textnormal{FACT}_{j}$ are given as in Algorithms~\ref{Algorithm1}--\ref{Algorithm2}, respectively. In addition, there exists some $C_{0}>0$ such that for all large $n$ and all large $z\ge 1$, 
 $$\mathbb{P}\left(\textnormal{FACT}_{j} \geq \frac{1}{2K}  (z-1) \right) \ge 1- C_{0} (|\kappa_{1}|\vee |\kappa_{2}|)^{-1}\sqrt{\frac{ K\log{z}}{n }} .$$
	\end{theorem}

The nonasymptotic results in	Theorem~\ref{theorem7} above for the FACT test complement the results of Theorem~\ref{theorem3} through the lens of power. It is seen that the magnitude of $\max_{1 \leq l \leq L}|\kappa_{l}|$ is crucial for having high selection power of the FACT test. In the remainder of this section, we provide a concrete example to illustrate the effectiveness of multiple transformation functions for power enhancement.

	\begin{condi}\label{model1}
	Assume that the nonparametric regression model is given by $Y = h(X_{j}) + H(\boldsymbol{X}_{-j}) + \varepsilon$,  where $h(\cdot)$ and $H(\cdot)$ are some measurable functions and $\varepsilon$ is the mean-zero model error that is independent of the random feature vector $\boldsymbol{X}$. In addition, assume that the distribution of feature vector $\boldsymbol{X}$ has a density function.
	\end{condi}
	
	\begin{proposition}\label{prop1}
		Assume that Condition~\ref{model1} holds, $0\le X_{l}\le 1$ for each $l\in\{1, \dots, p\}$, $\mathbb{E}|H(\boldsymbol{X}_{-j})|<\infty$, $h(\cdot)$ is monotonic, and the derivative of function $h(\cdot)$ is 
		integrable and bounded in absolute value. Then we have that 
		\[|\kappa_{1}| \ge \left(\inf_{x\in [0, 1]} \left|h'(x)\right| \right) \mathbb{E}\{\textnormal{Var}(X_{j}| \boldsymbol{X}_{-j})\}.\]
	\end{proposition}
	
	Proposition~\ref{prop1} above gives an example that when $\left(\inf_{x\in [0, 1]} \left|h'(x)\right| \right) \mathbb{E}\{\textnormal{Var}(X_{j}| \boldsymbol{X}_{-j})\} \gg (n/K)^{-1/2},$  
	the FACT test enjoys asymptotic power one in light of Theorem~\ref{theorem7} and \eqref{null.2}. The above condition rules out the pathological case when $X_j$ can be represented perfectly by a measurable function of $\boldsymbol{X}_{-j}$ almost surely. To motivate the need of using multiple transformations with $L\geq 2$, let us consider an example where Condition~\ref{model1} holds with $h(x) = (x - a)^{2}$, $a \in \mathbb{R}$, and $\boldsymbol{X}$ is uniformly distributed on $[0, 1]^{p}$. From  \eqref{new.eq.sec4.2.001} and some simple calculations, we can obtain that 
	\begin{equation}
	    \label{new.eq.sec4.2.002}
	    \kappa_{1} =  \frac{1}{12} - \frac{a}{6}, 
	\end{equation}   
 Thus, when $a$ is close to $0.5$, $\kappa_{1}$ is close to $0$, and hence the FACT test with $L=1$ does not have nontrivial power; the same can be concluded for the GCM test~\citep{shah2020hardness}. 
 The proposition below provides an example illustrating the advantage of using multiple transformations with $L\ge 2$ as suggested, for example, in \eqref{case.1}.

	\begin{proposition}\label{prop2}
		Assume that Condition~\ref{model1} holds for some $1 \leq j \leq p$ with $h(x) = a_{0} + \sum_{l=1}^{L} a_{l}x^{l}$, where $a_{l} \in \mathbb{R}$ and $\sum_{l=1}^{L}|a_{l}| > 0$. In addition, assume that $\mathbb{E}|H(\boldsymbol{X}_{-j})| <\infty$, $\boldsymbol{X}$ is uniformly distributed  on $[0, 1]^{p}$, and \eqref{case.1} holds. Then there exists some positive constant $c_{L}$ depending on $L$ such that
		$$
		\sum_{1=1}^{ L}|\kappa_{l}| \ge \frac{\sum_{l=1}^{L}|a_{l}|}{L}c_{L} >0.
		$$ 		 
		In particular, for $L=2$, we have 
		$|\kappa_{1}| + |\kappa_{2}| \ge 0.001\times (|a_{1}|+|a_{2}|).$ 
	\end{proposition}
	
The combined insights from Proposition~\ref{prop2} and Theorem~\ref{theorem7} above demonstrate that employing multiple proper transformation functions can boost the power to select features influencing the response via high-order polynomial terms. In particular, it is seen that for the setting considered in Proposition~\ref{prop2}, FACT with $L=2$ yields asymptotic power one as long as $|a_{1}|+|a_{2}|\gg \sqrt{K/n}$. Motivated by these results, we also consider the quadratic transformation for our practical version of FACT introduced in Section~\ref{fact.e.power}.  
 
\subsection{Stable FACT for large-scale multiple inference}\label{Sec4.3}

Through real data applications, we have observed that while the FACT inference helps control the FDR, the inherent randomness in random forests, stemming from column and row subsamplings, along with the randomness in sample splitting, can potentially reduce the reproducibility of the practical results. This is a recognized issue for methods employing subsampling or data-splitting techniques~\citep{meinshausen2009p}. To address such an issue, we introduce the stable FACT inference, which enhances the reproducibility by eliminating findings that may be susceptible to this intrinsic randomness.

Let e-values FACT$_{1}, \dots,$ FACT$_{p}$ be the outputs of Algorithm~\ref{Algorithm2}. We repeat Algorithm~\ref{Algorithm2} additional $B$ times for each $j$ and let the corresponding e-values be $\textnormal{FACT}_{j, 1}$, $\dots$, $\textnormal{FACT}_{j, B}$, where $B$ is a positive integer. Note that these repetitions of Algorithm~\ref{Algorithm2} are based on the same data, and conditional on the data, the randomness in these e-values comes from the intrinsic randomness in random forests and sample splitting. To put it differently, $\textnormal{FACT}_{j}, \textnormal{FACT}_{j, 1}, \dots, \textnormal{FACT}_{j, B}$ have the same distribution, and are conditionally independent given the data. We propose to consider the averaged e-value statistics $\widetilde{E}_{j}$'s to increase the stability of our inference results 
\begin{equation}
\widetilde{E}_{j} = B^{-1} \sum_{b=1}^B \textnormal{FACT}_{j, b}. 
\end{equation}
Then, using the e-BH procedure~\citep{wang2022false}, we obtain the set of selected features at target FDR level $\alpha$ as
\begin{equation}\label{reproducible.fact.1}
S^{\dagger}  = \{j: \widetilde{E}_{j} \ge  \widetilde{E}_{(k^{\dagger} ) } \},
\end{equation}
where $k^{\dagger} = \max\{ k : \frac{k \times \widetilde{E}_{(k)} }{ p } \ge \alpha^{-1} \}$ for some $\alpha\in (0, 1)$, and $\widetilde{E}_{(1 ) }$, $\dots$, $\widetilde{E}_{(p) }$ are the ordered statistics in descending order. This procedure is numerically more stable because of the aggregation across $B$ repetitions, with the caveat that the computation time is much higher for large $B$. 

We next build upon the above idea and introduce a more efficient implementation below. Define  $\widehat{S}  = \{j: \textnormal{FACT}_{j} \ge  \textnormal{FACT}_{(\widehat{k} ) } \}$ with $\widehat{k} = \max\{ k : \frac{k \times \textnormal{FACT}_{(k)} }{ p } \ge \alpha^{-1} \}$, where $\textnormal{FACT}_{(k )}$'s are the ordered statistics in descending order. Denote by 
\begin{equation}\label{new.fact.1}
 \widetilde{\textnormal{FACT}}_{j} = 
     \begin{cases}
         \textnormal{FACT}_{j} \ \textnormal{ if } j \not\in \widehat{S}, \\
         \min\left\{\textnormal{FACT}_{j}, B^{-1} \sum_{b=1}^B \textnormal{FACT}_{j, b} \right\} \ \textnormal{ if } j \in \widehat{S}.
     \end{cases}
 \end{equation}
For the ease of presentation, we refer to $\widehat{S}$ as the ``original'' FACT inference result. It is seen that the computation cost is much lower than that for ${S}^{\dagger}$. The stable FACT e-BH inference makes the discoveries
 \begin{equation}
     \label{reproducible.2}
     \widetilde{S}  = \{j: \widetilde{\textnormal{FACT}}_{j} \ge  \widetilde{\textnormal{FACT}}_{(\widetilde{k}) } \} \ \  (\textnormal{stable FACT inference}),
 \end{equation}
 where $\widetilde{k} = \max\{ k : \frac{k \times\widetilde{\textnormal{FACT}}_{(k)} }{ p } \ge \alpha^{-1} \}$.
The suggested stable FACT inference has nice statistical properties as shown in Proposition~\ref{prop5} below, where $\mathcal{H}_{0} \subset\{1, \dots, p\}$ is the index set of all null features.

\begin{proposition}
    \label{prop5}    
Assume that the e-values $\textnormal{FACT}_{j}$'s satisfy $\mathbb{E}(\textnormal{FACT}_{j})\le 1$ for each $j\in \mathcal{H}_{0}$. Then we have that $\mathbb{E} (\frac{\#(\widetilde{S}\cap \mathcal{H}_{0})}{\# \widetilde{S} \vee 1}) \le \alpha$, where $\alpha\in (0,1)$ is the target FDR level. In addition, it holds that $\widetilde{S} \subset S^{\dagger}$, where $S^{\dagger}$ is given in \eqref{reproducible.fact.1}.
\end{proposition}

Proposition~\ref{prop5} above shows that $\widetilde{S}$ still controls the FDR, at the cost of some power loss.  Our simulation study in Section~\ref{Sec5.4} later provides evidence that the power loss can be small.


\section{Simulation studies}
\label{Sec7}
	
In this section, we verify the theoretical properties of FACT and demonstrate its finite-sample performance through several simulation examples. We consider three data generating models
\begin{align}
        Y & = 5X_{1} + 2 X_{11} + \varepsilon,\label{Y1}\\        
        Y & = 5X_{1} + 10X_{2} + 20X_{6}^2 + 10\sin{(\pi X_{11} X_{12})} + \sqrt{5}\varepsilon, \label{Y5} \\
        Y & = 5X_{1}+\varepsilon,\label{Y3}
\end{align}
    where  $\varepsilon$ is the independent standard Gaussian model error. Models~\eqref{Y1} and \eqref{Y3} assume that $(X_{1}, \dots, X_{p})^T$ is a zero-mean multivariate Gaussian random vector with covariance matrix $\Sigma = (\lambda^{|k-l|})_{1\leq k,l\leq p}$ and $0\le \lambda<1$.  Model~\eqref{Y5} is a version of the Friedman regression function~\citep{friedman1991multivariate}, and the covariate distribution is chosen as a centered Gaussian copula with covariance matrix $\Sigma$. That is, $X_j = Z_j-0.5$ for $1\leq j\leq p$, with $(Z_1,\cdots, Z_p)$ having a joint CDF function $\Phi_{\Sigma}(\Phi^{-1}(z_1),\cdots, \Phi^{-1}(z_p))$, where $\Phi_{\Sigma}$ is the CDF of multivariate Gaussian with mean zero and covariance matrix $\Sigma$, and $\Phi^{-1}$ is the inverse of the univariate standard Gaussian CDF.  
    The Friedman regression function is frequently used for testing the selection power for the linear main effect, quadratic effect, and interaction effect. We note that  model~\eqref{Y5} differs from the original  Friedman regression function in that Gaussian copula instead of the uniform distribution is used; this is because feature correlation is a well-known contributor to the bias issue of existing random forests feature importance measures, but the uniform distribution does not introduce any correlation among features. 
    
    We consider different dimensionalities $p\in \{15, 30, 150, 200 \}$. In our numerical studies, all test statistics given in Algorithms~\ref{Algorithm1}--\ref{Algorithm2} are calculated based on the OOB random forests prediction as detailed in Section~\ref{Sec.new3}.

		\begin{figure}[t]
		\centering
		\captionsetup{width=.3\linewidth}
		\subfloat[F$_{2}^{(1)}$ with $K=1$ and $\lambda = 0.5$.]{\includegraphics[width=5.1cm]{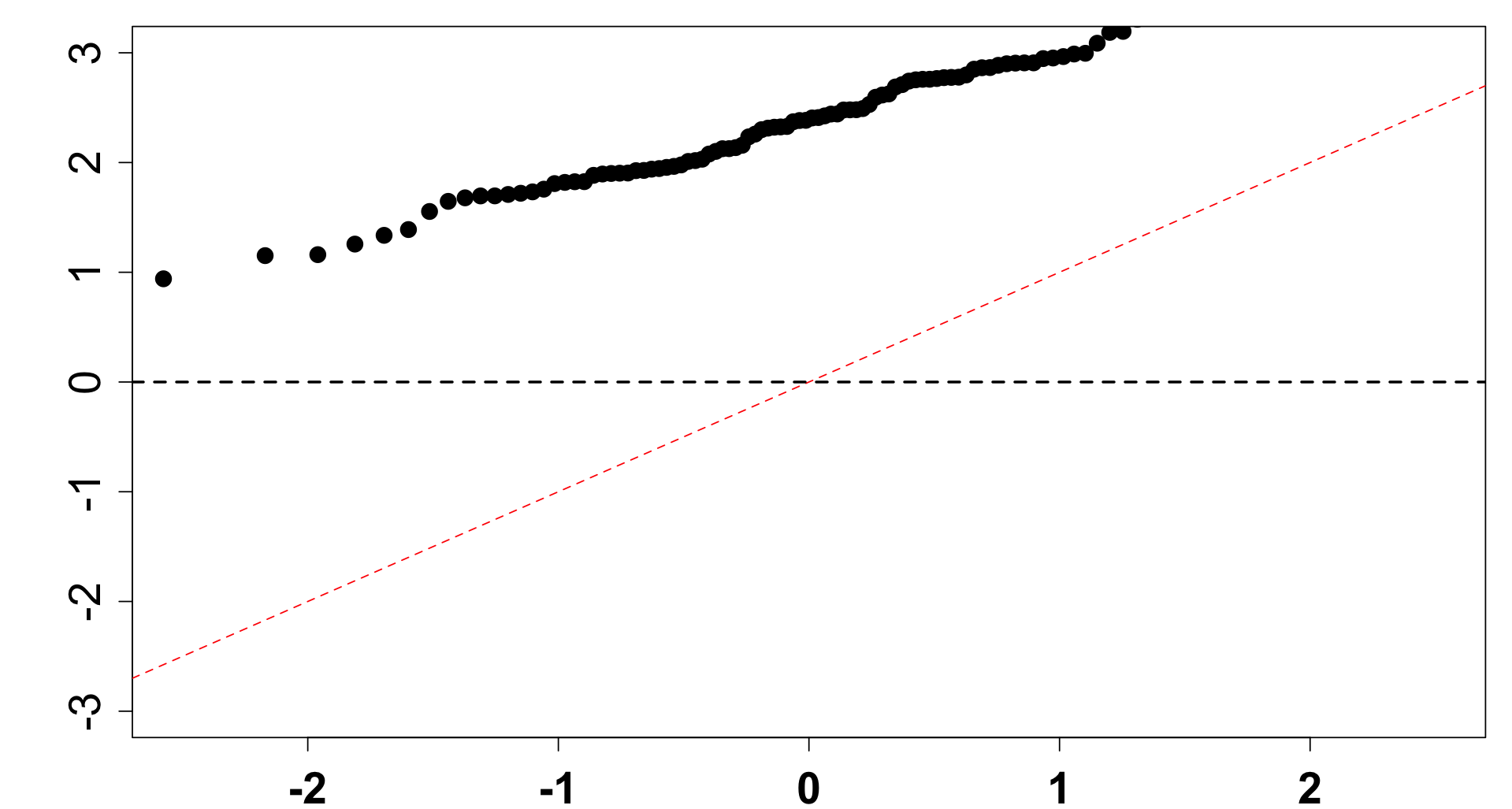}}
		\subfloat[F$_{2}^{(1)}$ with $K=3$ and $\lambda = 0.5$.]{\includegraphics[width=5.1cm]{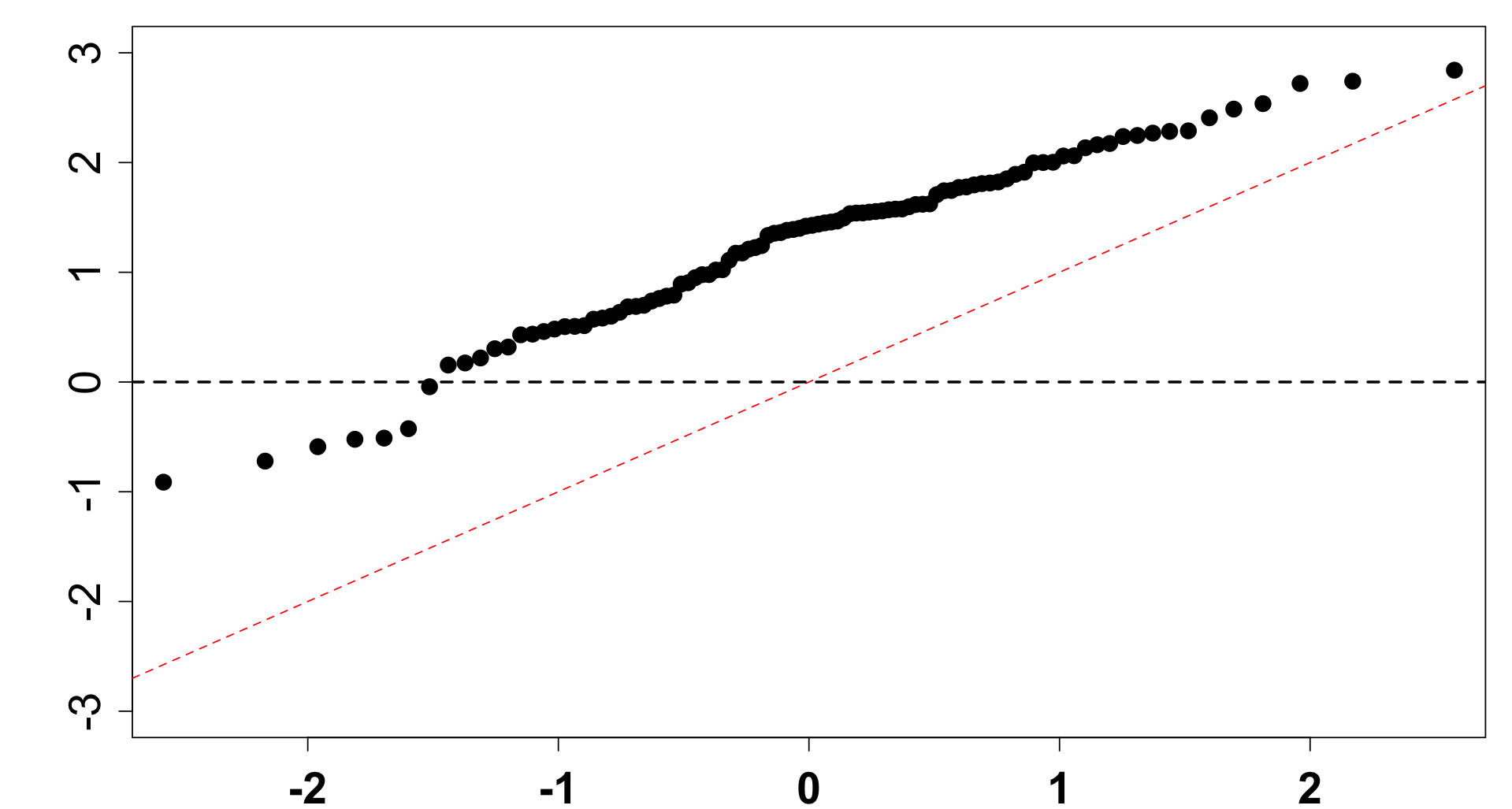}}
  \subfloat[F$_{2}^{(1)}$ with $K=10$ and $\lambda = 0.5$.]{\includegraphics[width=5.1cm]{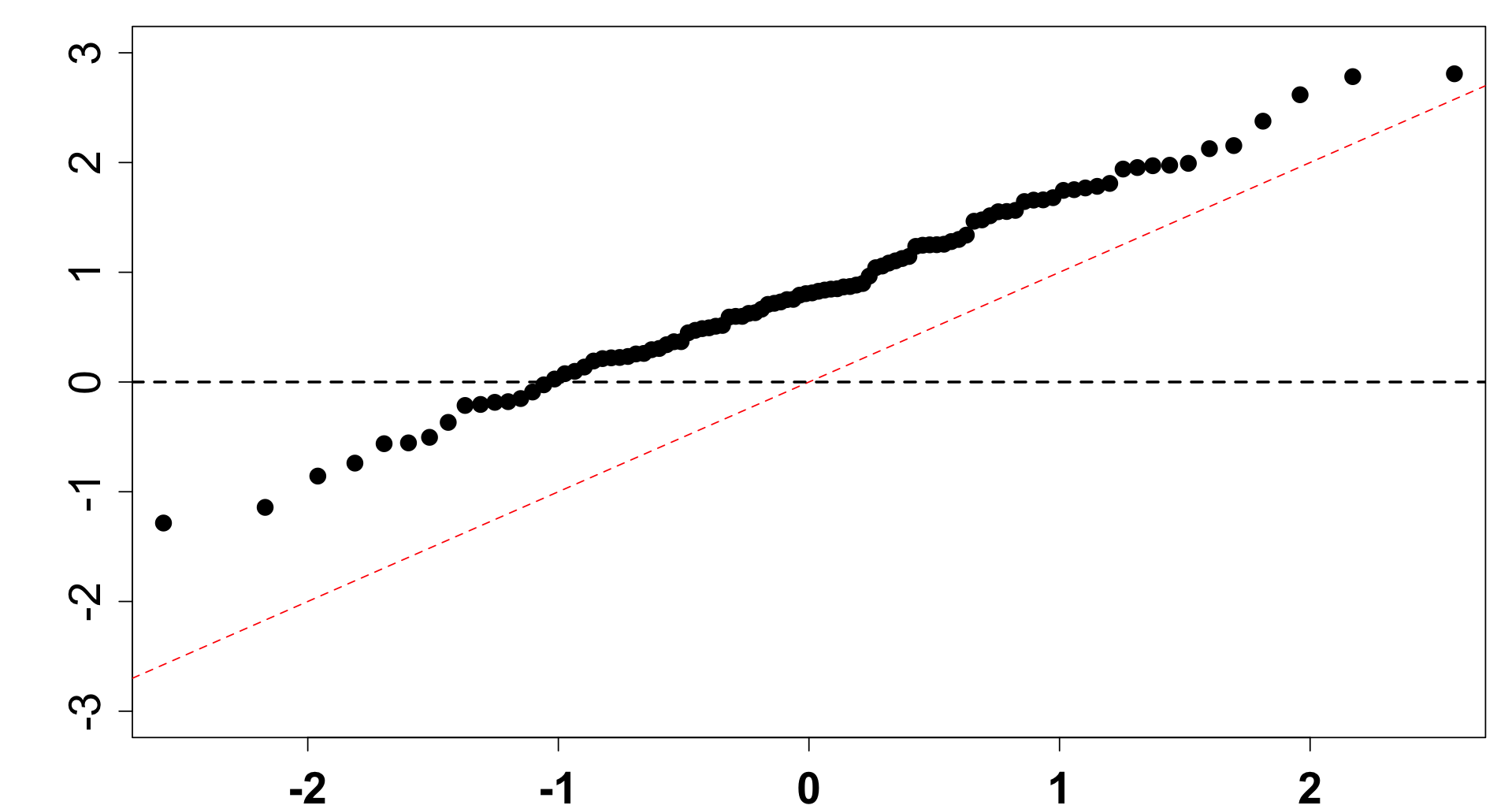}}

  \subfloat[F$_{12}^{(1)}$ with $K=1$ and $\lambda = 0.5$.]{\includegraphics[width=5.1cm]{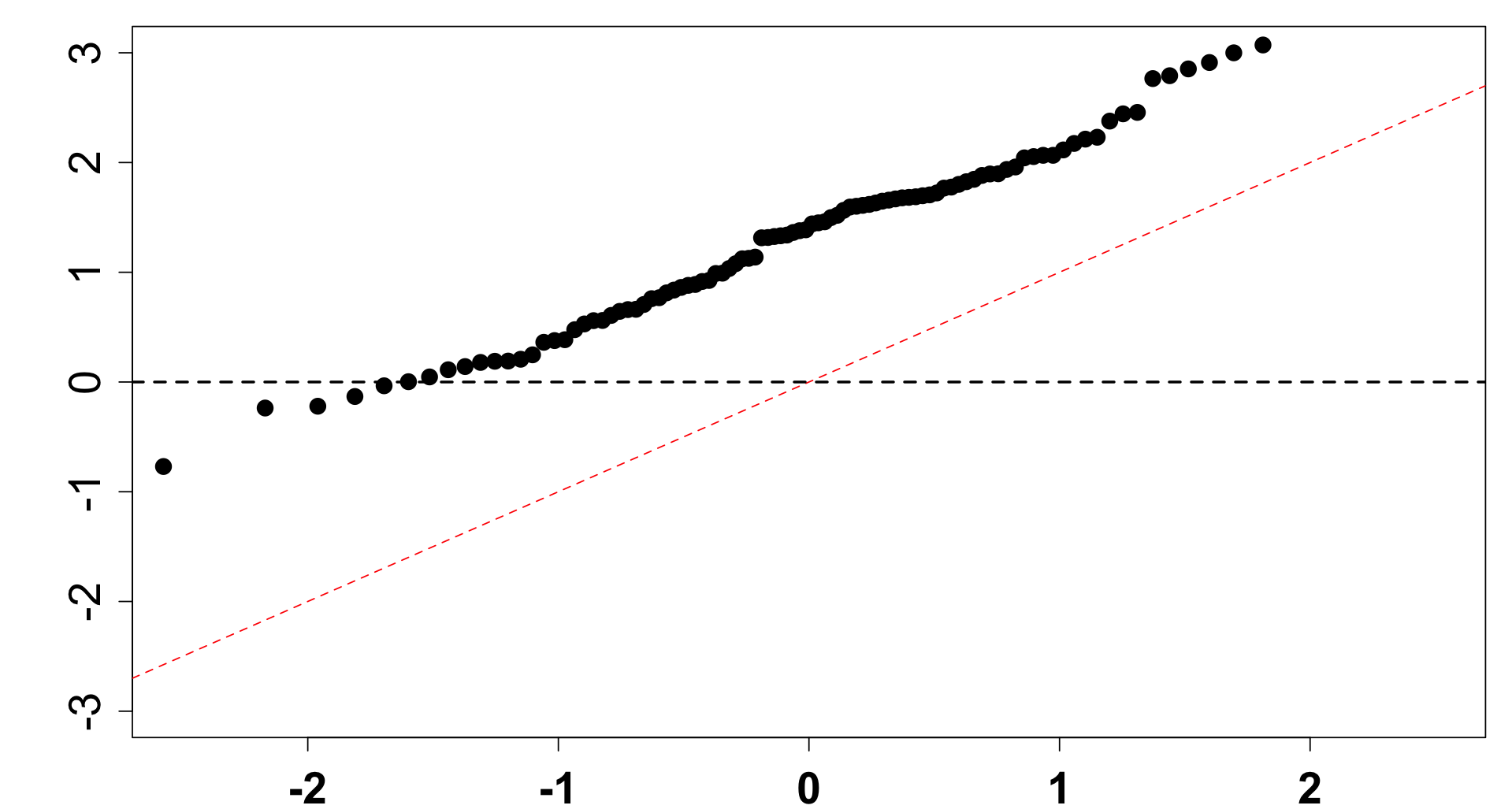}}
  \subfloat[F$_{12}^{(1)}$ with $K=3$ and $\lambda = 0.5$.]{\includegraphics[width=5.1cm]{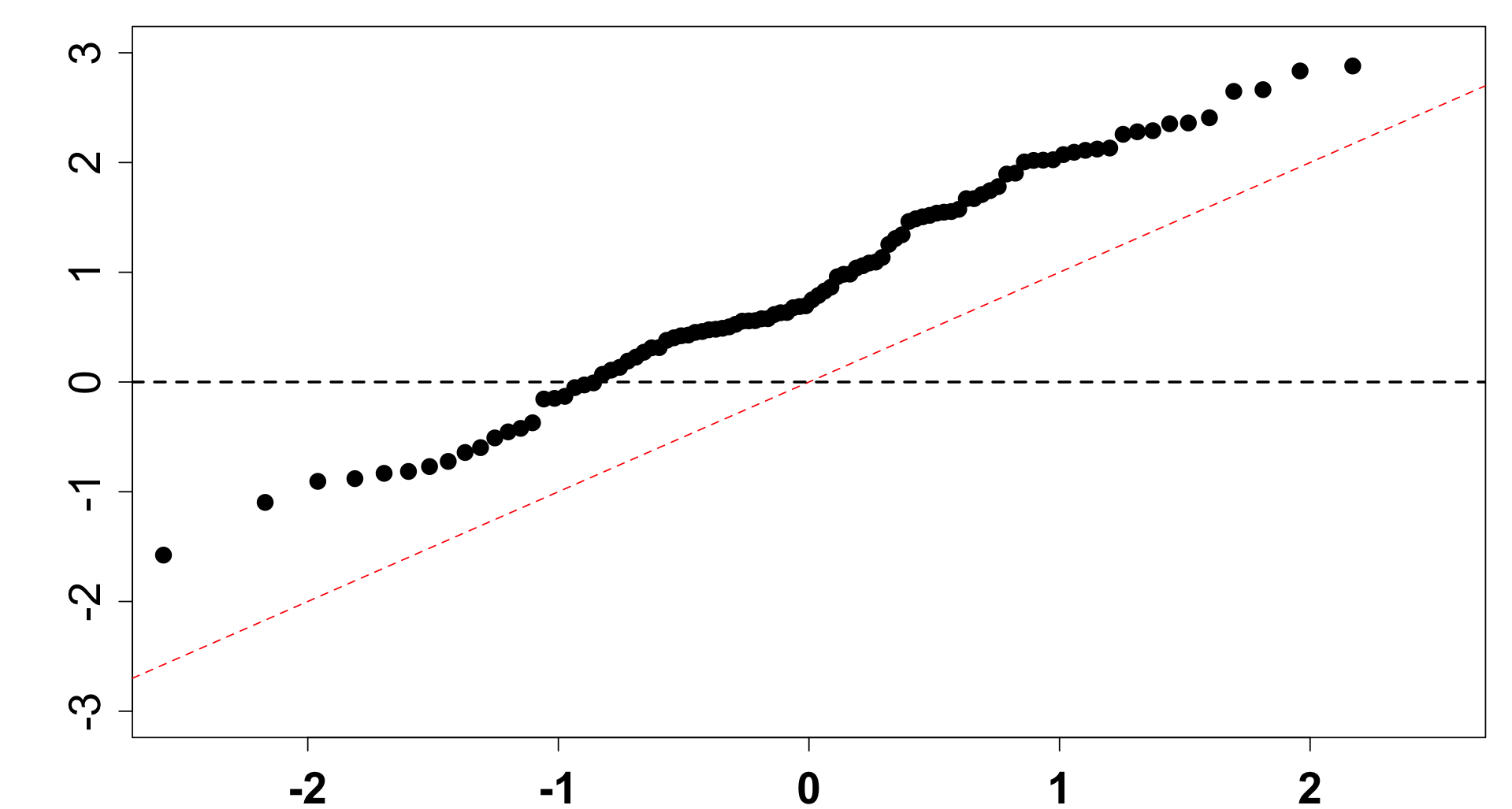}}
  \subfloat[F$_{12}^{(1)}$ with $K=10$ and $\lambda = 0.5$.]{\includegraphics[width=5.1cm]{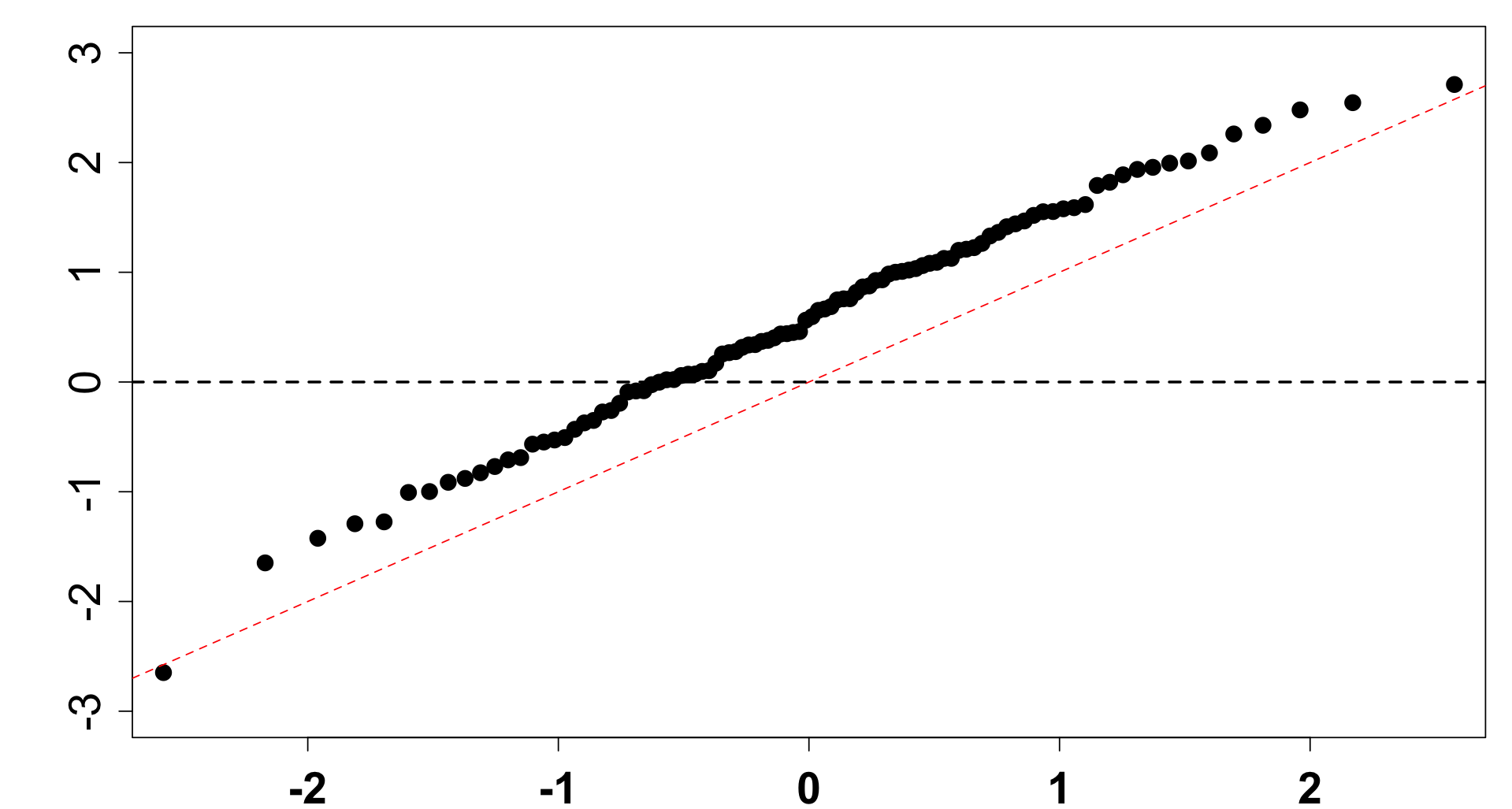}}
		
		\captionsetup{width=1.\linewidth}
		\caption{Each Q-Q plot for the FACT test statistics against the standard Gaussian distribution from model~\eqref{Y1} in Section \ref{Sec7.1} consists of $100$ points, each of which is based on a sample with $n = 500$, $p=15$. Panels a) and d) are the results for the GCM. The red lines represent the 45-degree lines passing through the origin.}
		\label{fig:2}
	\end{figure}

    \subsection{Asymptotic normality} \label{Sec7.1}
    We examine the empirical distributions of the GCM and FACT statistics for the null features using the quantile-quantile (Q-Q) plots. 
	To this end, we generate $100$ data sets, each with sample size $n =500$ and feature dimensionality $p = 15$ from  model \eqref{Y1} with the choices of $\lambda=0.5$. We then calculate the test statistics $F_{2}^{(1)}$ given in \eqref{E5} with $K\in \{1, 3, 10\}$ and $g_{1}(x) = g_{2}(x) = x$; note that the statistics $F_{2}^{(1)}$ in panels (a) and (d) are the GCM test statistics. The resulting Q-Q plots are presented in Figure~\ref{fig:2} with the 45-degree red lines passing through the origin.

  It is seen from panels (a) and (d) of Figure~\ref{fig:2} that the empirical distribution of GCM deviates from the standard Gaussian due to the bias, and the deviation is severe even with medium correlation level $\lambda = 0.5$. We also demonstrate in the other panels that FACT alleviates the bias issue via the use of imbalanced samples $K \in \{3, 10\}$.

 \begin{table}[t]
		\begin{center}
			{
				\begin{tabular}[t]{ l|l|l|c|c|c|c|c|c|c|c}
					\hline &\multirow{2}{*}{$(n, p, \lambda)$ }&\multirow{2}{*}{ $\alpha$}&\multicolumn{5}{c|}{Relevant features} &\multicolumn{3}{c}{Null features}\\
					\cline{4-11}
					&&& \multicolumn{1}{c}{$X_{1}$} &\multicolumn{1}{c}{$X_{2}$} & \multicolumn{1}{c}{$X_{6}$}& \multicolumn{1}{c}{$X_{11}$}& \multicolumn{1}{c|}{$X_{12}$} &  
					\multicolumn{1}{c}{$X_{3}$} & \multicolumn{1}{c}{$X_{7}$}& \multicolumn{1}{c}{$X_{13}$}    \\ \hline
					\multirow{3}{*}{\textnormal{I}}&\multirow{3}{*}{$(150, 30, 0.3)$} &\multirow{1}{*}{$ 0.1\phantom{00}$} & 0.88 & 1.00 & 0.66  &0.11 & 0.05 & 0.00 &0.00 & 0.00   \\
					&&\multirow{1}{*}{$0.05\phantom{0}$} & 0.66 & 1.00 & 0.44  & 0.04 & 0.03 & 0.00 & 0.00 & 0.00   \\
					&& $0.025\phantom{}$ & 0.48 & 1.00 & 0.33  &0.00 &0.00 &0.00 &0.00 &0.00 \\ \hline 
					\multirow{3}{*}{\textnormal{II}}&\multirow{3}{*}{$( 150, 30, 0.7)$} &\multirow{1}{*}{$ 0.1\phantom{00}$} & 0.72 & 1.00 & 0.72 & 0.48 & 0.52 &  0.01 &  0.00  & 0.02   \\ 
					&&\multirow{1}{*}{$0.05\phantom{0}$} & 0.55 & 0.99 & 0.50 & 0.34 & 0.28 & 0.00 & 0.00 & 0.00  \\
					&& $0.025\phantom{}$ & 0.36 & 0.97 & 0.28 & 0.23 &0.18 & 0.00 & 0.00 & 0.00 \\ \hline
					\multirow{3}{*}{\textnormal{III}}&\multirow{3}{*}{$( 250, 30, 0.3)$} &\multirow{1}{*}{$ 0.1\phantom{00}$} & 0.99 & 1.00 & 0.94  & 0.13 & 0.27 & 0.00 & 0.01 & 0.00   \\ 
					&&\multirow{1}{*}{$0.05\phantom{0}$} & 0.95 & 1.00 & 0.87  & 0.07 & 0.09 & 0.00 & 0.01 & 0.00   \\  
					&& $0.025\phantom{}$ & 0.86 & 1.00 & 0.81  & 0.03 & 0.03 & 0.00 & 0.00 & 0.00  \\ \hline
     
                \multirow{3}{*}{\textnormal{IV}}&\multirow{3}{*}{$( 250, 30, 0.7)$} &\multirow{1}{*}{$ 0.1\phantom{00}$} & 0.99 & 1.00 & 0.95 & 0.89  & 0.89  & 0.00 & 0.02 &0.02   \\ 
					&&\multirow{1}{*}{$0.05\phantom{0}$} & 0.92 & 1.00 & 0.86 & 0.74  & 0.72  & 0.00 & 0.01 & 0.01  \\  
					&& $0.025\phantom{}$ & 0.82 & 1.00 & 0.71 & 0.56  & 0.55  & 0.00 & 0.00 & 0.01 \\ \hline    \multirow{3}{*}{\textnormal{V}}&\multirow{3}{*}{$( 350, 150, 0.7)$} &\multirow{1}{*}{$ 0.1\phantom{00}$} & 1.00 & 1.00 & 0.99  & 0.98 & 0.99 & 0.00 & 0.02 & 0.03   \\ 
					&&\multirow{1}{*}{$0.05\phantom{0}$} & 0.99 & 1.00 & 0.96  & 0.97 & 0.98 & 0.00 & 0.01 & 0.00   \\  
					&& $0.025\phantom{}$ & 0.94 & 1.00 & 0.94  & 0.93 & 0.92 & 0.00 & 0.00 & 0.00  \\ \hline
					\multicolumn{10}{c}{ }
			\end{tabular} }

			\caption{The empirical size and power of the FACT test under model~\eqref{Y5} with Gaussian copula at each significance level $\alpha\in\{0.1, 0.05, 0.025\}$ over $100$ simulation repetitions.} \label{tab:4}
		\end{center} 
	\end{table}	
	\begin{table}[t]
		\begin{center}
			{
				\begin{tabular}[t]{ l|l|l|c|c|c|c|c|c|c|c}
					\hline &\multirow{2}{*}{$(n, p, \lambda)$ }&\multirow{2}{*}{ $\alpha$}&\multicolumn{5}{c|}{Relevant features} &\multicolumn{3}{c}{Null features}\\
					\cline{4-11}
					&&& \multicolumn{1}{c}{$X_{1}$} &\multicolumn{1}{c}{$X_{2}$} & \multicolumn{1}{c}{$X_{6}$}& \multicolumn{1}{c}{$X_{11}$}& \multicolumn{1}{c|}{$X_{12}$} &  
					\multicolumn{1}{c}{$X_{3}$} & \multicolumn{1}{c}{$X_{7}$}& \multicolumn{1}{c}{$X_{13}$}    \\ \hline
					\multirow{3}{*}{\textnormal{I}}&\multirow{3}{*}{$(150, 30, 0.3)$} &\multirow{1}{*}{$ 0.1\phantom{00}$} & 1.00 & 1.00 & 0.13  &0.12 & 0.09 & 0.15 &0.08 & 0.12   \\
					&&\multirow{1}{*}{$0.05\phantom{0}$} & 0.99 & 1.00 & 0.07  & 0.07 & 0.02 & 0.06 & 0.02 & 0.01   \\
					&& $0.025\phantom{}$ & 0.99 & 1.00 & 0.03  &0.05 &0.00 &0.03 &0.00 &0.00 \\ \hline
					\multirow{3}{*}{\textnormal{II}}&\multirow{3}{*}{$( 150, 30, 0.7)$} &\multirow{1}{*}{$ 0.1\phantom{00}$} & 1.00 &1.00 & 0.11  &0.06 &0.02 &0.15 &0.04 &0.04  \\  
					&&\multirow{1}{*}{$0.05\phantom{0}$} & 1.00 & 1.00 & 0.03  & 0.03 & 0.01 & 0.09 & 0.01 & 0.01 \\  
					&& $0.025\phantom{}$ & 1.00 & 1.00 & 0.02 & 0.02 & 0.00 & 0.03 & 0.01 & 0.01 \\ \hline  
					\multirow{3}{*}{\textnormal{III}}&\multirow{3}{*}{$( 250, 30, 0.3)$} &\multirow{1}{*}{$ 0.1\phantom{00}$} & 1.00 & 1.00 & 0.09  & 0.11 & 0.13 & 0.16 & 0.05 & 0.06   \\ 
					&&\multirow{1}{*}{$0.05\phantom{0}$} & 1.00 & 1.00 & 0.05  & 0.03 & 0.08 & 0.11 & 0.02 & 0.01   \\  
					&& $0.025\phantom{}$ & 1.00 & 1.00 & 0.02  & 0.01 & 0.03 & 0.05 & 0.01 &0.00  \\ \hline
     
     \multirow{3}{*}{\textnormal{IV}}&\multirow{3}{*}{$( 250, 30, 0.7)$} &\multirow{1}{*}{$ 0.1\phantom{00}$} & 1.00 & 1.00 & 0.12  & 0.11 & 0.12 & 0.16 & 0.07 & 0.09   \\ 
					&&\multirow{1}{*}{$0.05\phantom{0}$} & 1.00 & 1.00 & 0.08  & 0.04 & 0.07 & 0.13 & 0.01 & 0.05   \\  
					&& $0.025\phantom{}$ & 1.00 & 1.00 & 0.08  & 0.04 & 0.04 & 0.05 & 0.01 &0.02  \\ \hline
     \multirow{3}{*}{\textnormal{V}}&\multirow{3}{*}{$( 350, 150, 0.7)$} &\multirow{1}{*}{$ 0.1\phantom{00}$} & 1.00 & 1.00 & 0.10  & 0.08 & 0.07 & 0.32 & 0.09 & 0.03   \\ 
					&&\multirow{1}{*}{$0.05\phantom{0}$} & 1.00 & 1.00 & 0.05  & 0.04 & 0.04 & 0.20 & 0.03 & 0.01   \\  
					&& $0.025\phantom{}$ & 1.00 & 1.00 & 0.00  & 0.03 & 0.01 & 0.15 & 0.01 & 0.00  \\ \hline
					\multicolumn{10}{c}{ }
			\end{tabular} }
			\caption{The empirical size and power of the GCM test under model~\eqref{Y5} with Gaussian copula at each significance level $\alpha\in\{0.1, 0.05, 0.025\}$ over $100$ simulation repetitions.} \label{tab:4.gcm}
		\end{center}
	\end{table}

 
	\subsection{Hypothesis testing size and power} \label{Sec7.2}
	
	We now investigate the empirical performance of the FACT test, FACT$_{j}$, with $K = L=2$, $g_{11}(x) = g_{12}(x) = g_{21}(x) =\tanh(x)$, and $g_{22}(x) = (\tanh(x))^2$ introduced in Algorithm~\ref{Algorithm2}
	in terms of the size and power. We consider different significance levels $\alpha \in \{0.1, 0.05, 0.025\}$. For testing the null hypothesis \eqref{null3}, we calculate the corresponding FACT e-values, and reject the null hypothesis if $ \textnormal{FACT}_{j} \ge \alpha^{-1}$, as discussed after Theorem~\ref{theorem3}. 

For each feature $X_j$ with $j\in \{1,2, 6, 11, 12, 3, 7, 13\}$, the empirical rejection rates over $100$ simulation repetitions are reported in Table~\ref{tab:4}, where the data is generated from the nonparametric model \eqref{Y5}. Specifically, we consider five cases of $(n, p, \lambda)$: I) $(n, p, \lambda) = (150, 30, 0.3)$, II) $(n, p, \lambda) = (150, 30, 0.7)$, III) $(n, p, \lambda) = (250, 30, 0.3)$, IV) $(n, p, \lambda) = (250, 30, 0.7)$, and V) $(n, p, \lambda) = (350, 150, 0.7)$. As a comparison, in Table~\ref{tab:4.gcm} we present the corresponding results of GCM, which is F$_{j}^{(1)}$ defined in \eqref{E5} with $K=1$ and transformation functions $g_{1}(x) = g_{2}(x) = x$; see Section~\ref{fact-algorithm} or Section~\ref{Sec7.1} for details of GCM.

Our results in Tables~\ref{tab:4}--\ref{tab:4.gcm} show that the FACT controls the type I error below the target level in all instances, but the GCM is seriously biased when testing $X_{3}$; a potential explanation is that $X_3$ is highly correlated with the linear components $(X_{1}, X_{2})$ in model~\eqref{Y5}. In addition, although the GCM controls the type I error for $X_{7}$ and $X_{13}$, which are highly correlated with the quadratic component $X_{6}$ and the interaction components $(X_{11}, X_{12})$, respectively, it does not have selection power for $(X_{6}, X_{11}, X_{12})$ in model~\eqref{Y5}. A potential explanation is that the dependency is nonlinear, as we illustrated in Section~\ref{Sec4.2}.  In contrast, the FACT can identify relevant features $(X_{6}, X_{11}, X_{12})$ much more effectively in all cases except for the interaction components $(X_{11}, X_{12})$ in cases I and III. The power results demonstrate the improved selection power of FACT due to the use of multiple transformation functions. Indeed, a careful examination of the result reveals that the power of FACT for $X_{1}$ and $X_{2}$ mainly comes from the e-value statistic $e_{j}(\tanh{(x)}, \tanh{(x)})$, while the power for testing $(X_{6}, X_{11}, X_{12})$ is mainly from  the e-value statistic $e_{j}(\tanh{(x)}, (\tanh{(x)} )^2)$. These results provide evidence supporting the use of multiple transformations for capturing the nonlinear dependency in the data. 


Cases II, IV, and V in Table~\ref{tab:4} show that the empirical selection power of FACT for $(X_{1}, X_{2}, X_{6})$ increases quickly as the sample size increases, regardless of the correlation level. On the other hand, the selection power of FACT for the interaction components $(X_{11}, X_{12})$ depends on the covariate correlation. The selection power only increases with the sample size when the correlation is high. We observe empirically the interesting phenomenon that high correlation contributes favorably toward the successful selection of interaction variables. Inspired by our power analysis in Section \ref{Sec4.2}, the power is jointly determined by the signal strength $\kappa_l$ and the random forests prediction accuracy $B_1$ and $B_2$. Since the covariates correlation can affect both quantities in rather complicated ways in general interaction models, and  the variable transformations add to the complication, we defer the detailed investigation on the effect of covariates correlation on power to a future study.  



\subsection{
Comparisons with MDI, MDA, and CPI} \label{Sec7.3}

We now provide details on the results in Table \ref{tab:5} in the Introduction. Consider the first four settings of model \eqref{Y1} in Section \ref{Sec7.2}. 
To calculate the MDI and MDA measures, we employ the R package \texttt{randomForest}~\citep{rf}, while for the calculation of the CPI measure \citep{debeer2020conditional, strobl2008conditional}, we use the R package \texttt{permimp}~\citep{debeer2021package}. The computation is done with the default configurations of those R packages. Meanwhile, the FACT statistics in Algorithm~\ref{Algorithm2} are calculated with $K = L=2$, $g_{11}(x) = g_{12}(x) = g_{21}(x) =\tanh(x)$, and $g_{22}(x) = (\tanh(x))^2$. We provide in Table~\ref{tab:5} the simulation results for examining the spurious effects of different random forests feature importance measures and the FACT statistics with respect to the null feature $X_{2}$, which is correlated with the strong linear component $X_{1}$ in model \eqref{Y1}. Specifically, each entry of Table~\ref{tab:5} stands for the fraction of times (out of 100 simulation repetitions) when the feature importance measure of the null spurious feature $X_{2}$ exceeds that of relevant feature $X_{11}$. A larger entry in Table~\ref{tab:5} suggests a stronger spurious effect.

Table~\ref{tab:5} unveils several interesting phenomena on the spurious effects of different random forests feature importance measures. First, we see from Table~\ref{tab:5} that the importance of the null spurious feature $X_{2}$ dominates frequently that of the relevant feature $X_{11}$  for all three feature importance measures MDI, MDA, and CPI across cases II--III, due to the high correlation between the null feature $X_{2}$ and the relevant feature $X_{1}$. We also see that the use of the CPI measure alleviates the spurious effects to certain extent compared to the MDI and MDA measures. In sharp contrast, the feature significance measure of the FACT statistic suppresses the spurious effects satisfactorily across all cases I--III.

\subsection{Stable FACT inference for multiple comparisons}\label{Sec5.4}

Proposition~\ref{prop5} has shown that \eqref{reproducible.2} yields stable selection results with controlled FDR, at the cost of some power loss. Here, we use a simulation study to verify these theoretical results. We will also demonstrate that the power loss is not severe.  To reduce the computation cost, we use the original FACT inference (see Section~\ref{Sec4.3} for details) as a reference point when we examine the power of the stable FACT. We consider the scenarios when only a few observations are available.

We consider the data generating model~\eqref{Y3} with $n\in \{24, 48\}$, $p=200$, and $\lambda \in \{0, 0.7\}$, and set $B = 100$ in \eqref{new.fact.1}. The FACT statistics are calculated using Algorithm~\ref{Algorithm2} with $K = L=2$, $g_{11}(x) = g_{12}(x) = g_{21}(x) =\tanh(x)$, and $g_{22}(x) = (\tanh(x))^2$. The detailed implementation of the stable FACT can be found in \eqref{reproducible.2}.  Table~\ref{tab:reproducible.1} summarizes the empirical size and power of our stable FACT inference procedure, with FDR level set at $\alpha = 0.2$. We note that the stable FACT is slightly more conservative in terms of the FDR control, and its power is marginally lower compared to that of the original FACT. These outcomes align well with Proposition~\ref{prop5}.
Moreover, in Table~\ref{tab:reproducible.1}, we observe a notable improvement in selection power with a slight increase in the sample size (i.e., from $n=24$ to $n=48$), both for the original and stable FACT. In summary, the results in Table~\ref{tab:reproducible.1} suggest that the stable FACT offers a balanced compromise between the original FACT and the fully aggregated FACT as defined in \eqref{reproducible.fact.1}.

\begin{table}
{\small
		\begin{tabular}[t]{cc} 
			\begin{tabular}[t]{ c|cc| cc}	\hline	
    $n=24 $&\multicolumn{2}{c|}{ FACT} & \multicolumn{2}{c}{ Stable FACT~\eqref{new.fact.1}}  
   \\ 
			$\lambda$	&\multicolumn{1}{c}{ Power} & \multicolumn{1}{c|}{FDR}& \multicolumn{1}{c}{ Power} & \multicolumn{1}{c}{ FDR} \\
				\hline					
				$0$ &0.48 & 0.06  & 0.45 & 0.005 \\
				\hline
				$0.7$ &0.57 & 0.07  & 0.51 & 0  \\
				\hline				
			\end{tabular}	
			\ \ 
		\end{tabular}
  \begin{tabular}[t]{cc} 
			\begin{tabular}[t]{ c|cc| cc}	\hline	
    $n=48 $&\multicolumn{2}{c|}{ FACT} & \multicolumn{2}{c}{ Stable FACT~\eqref{new.fact.1}}  
   \\ 
			$\lambda$	&\multicolumn{1}{c}{ Power} & \multicolumn{1}{c|}{FDR}& \multicolumn{1}{c}{ Power} & \multicolumn{1}{c}{ FDR} \\
				\hline					
				$0$ & 0.95 & 0  & 0.94 & 0 \\
				\hline
				$0.7$ & 0.85 & 0.01  & 0.82 & 0  \\
				\hline				
			\end{tabular}	
			\ \ 
		\end{tabular}
		\captionof{table}{The empirical FDR and power are calculated based on the formulas given in Section~\ref{Sec5.4}, where $\lambda$ represents the correlation level among the covariates. }\label{tab:reproducible.1}
	}
	\end{table}

\section{Real data application} \label{Sec8}

\begin{figure}[tp]
    \centering
    \includegraphics[width=11cm]{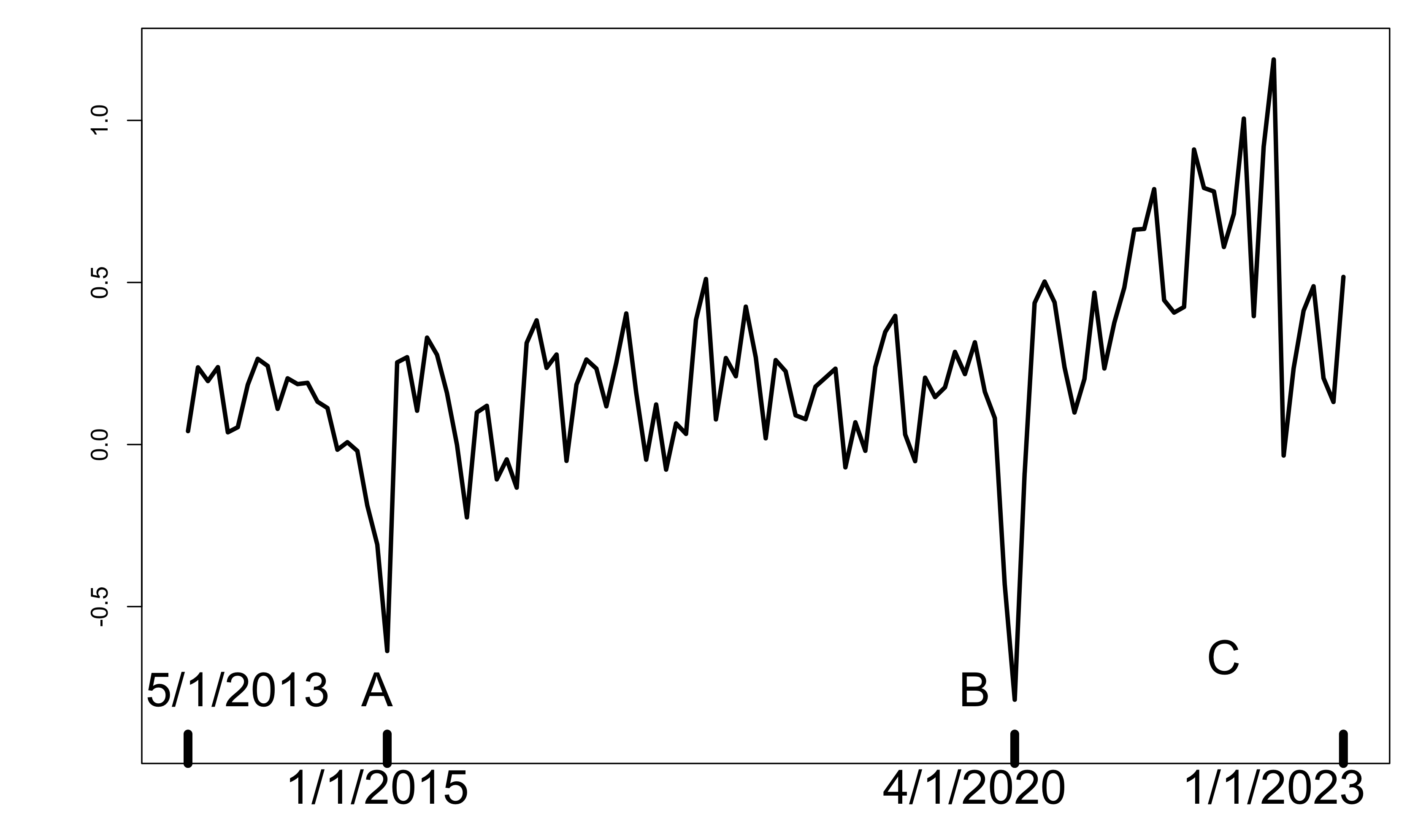}
    \caption{The U.S. inflation from May 2013 to January 2023.}
    \label{fig:inflation}
\end{figure}

In this section, we analyze the temporal relations between the U.S. inflation and other $126$ major time series covariates of the U.S. economy such as the exchange rates, housing prices, and industrial prices, from May 2013 to January 2023 using the monthly FRED-MD data during this period~\citep{mccracken2016fred}. The inflation at month $t$ is defined as  
$$\textnormal{Inflation}_{t} = \left(\frac{\textnormal{CPI}_{t} - \textnormal{CPI}_{t-1}}{\textnormal{CPI}_{t-1}} \times 100\right) \%, $$
where CPI$_{t}$ represents the consumer price index for all goods at month $t$ (the FRED-MD code of this series is CPIAUCSL). Figure~\ref{fig:inflation} shows the versatile time-varying patterns of the inflation series across time, with three notable time points indicating obvious pattern changes. To ease the presentation, we name the events at January 2015 and April 2020 as A and B, respectively, and name the period  of economic recovery in the post-COVID-19 era and Russia--Ukraine war as C, as shown in Figure~\ref{fig:inflation}. Because of these obvious pattern changes, it is unsuitable to study the dependency of inflation on other time series variables based on the entire time period.

We investigate this problem using the FACT for two main reasons. First, identifying important time series economic variables that can affect the inflation has been an active research problem with a long history~\citep{king1995temporal, crump2022unemployment}, due to the importance of inflation. Second, the nonstationarity of inflation time series motivates the use of rolling windows, yielding a small sample size for each window period. Hence, the FACT inference framework is suitable here, as we have illustrated in Section~\ref{Sec5.4} that stable FACT can control FDR with reasonable power even with sample size $n=24$, equivalent to a two-year rolling window period with monthly data.

\begin{figure}[tp]
    \centering
    \subfloat[The red curve is EXUSUKx.]{
    \includegraphics[width=10.cm]{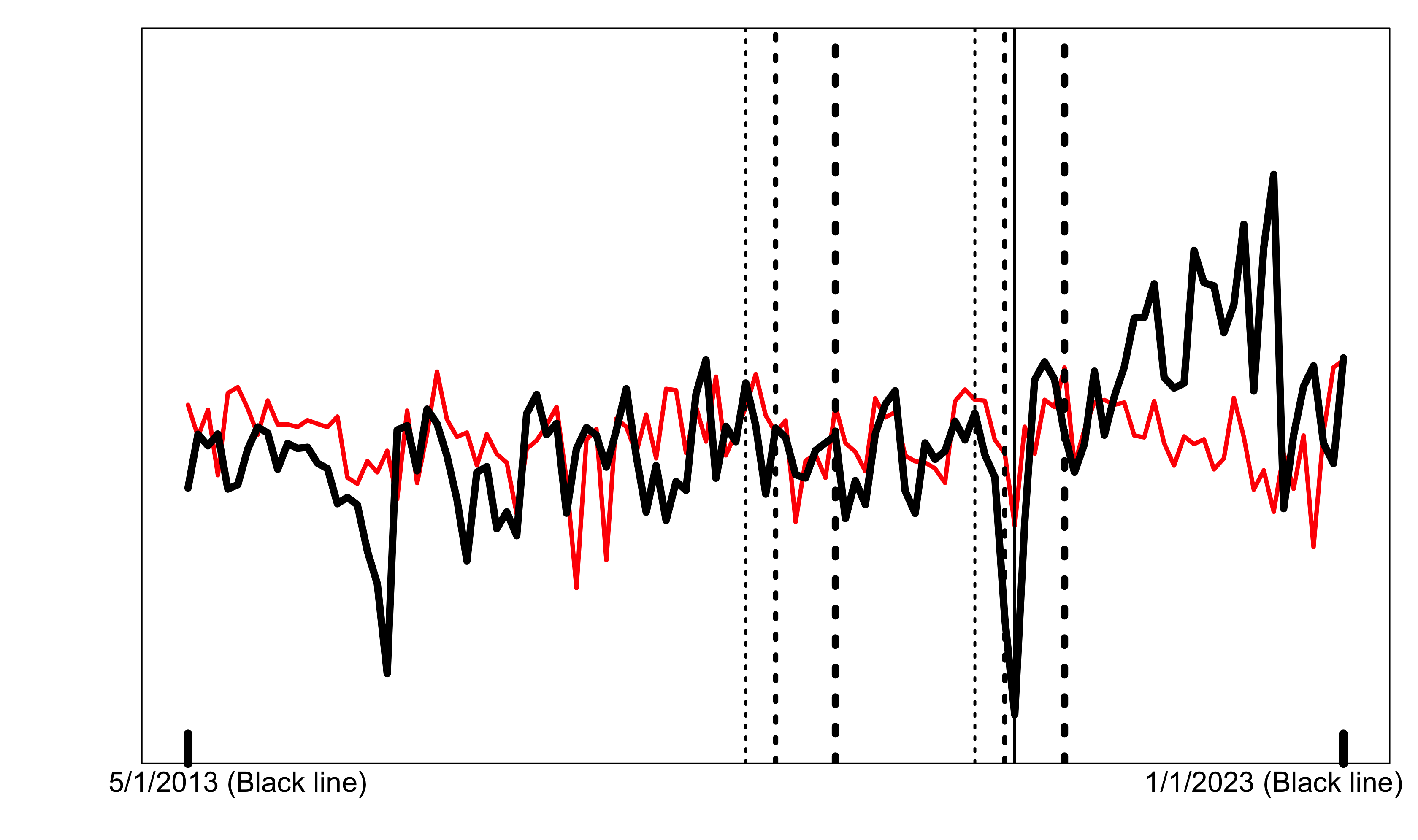}}%

    \subfloat[The red curve is CUUR0000SAD.]{\includegraphics[width=10.cm]{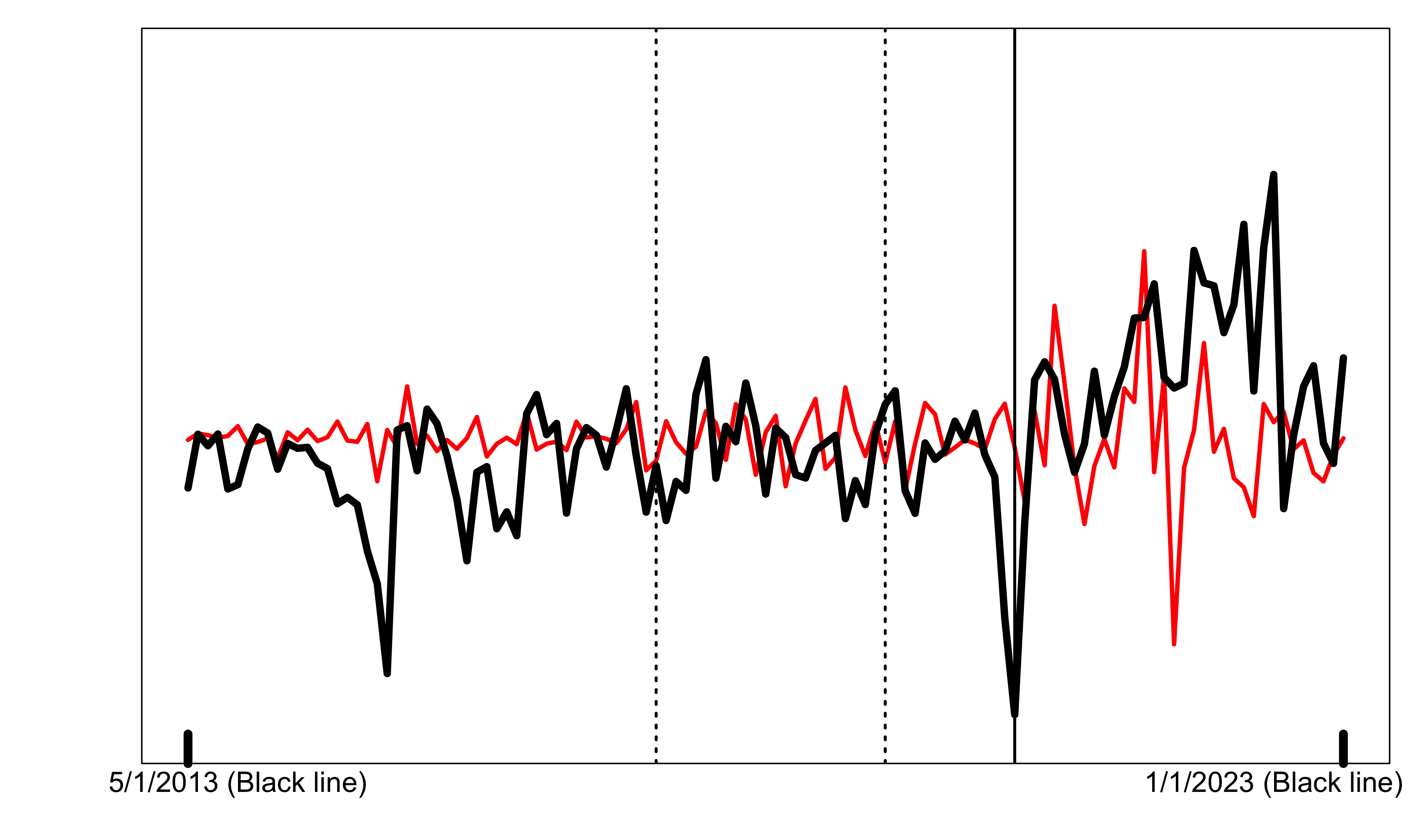}}%
    
    \subfloat[The red curve is IPNMAT.]{\includegraphics[width=10.cm]{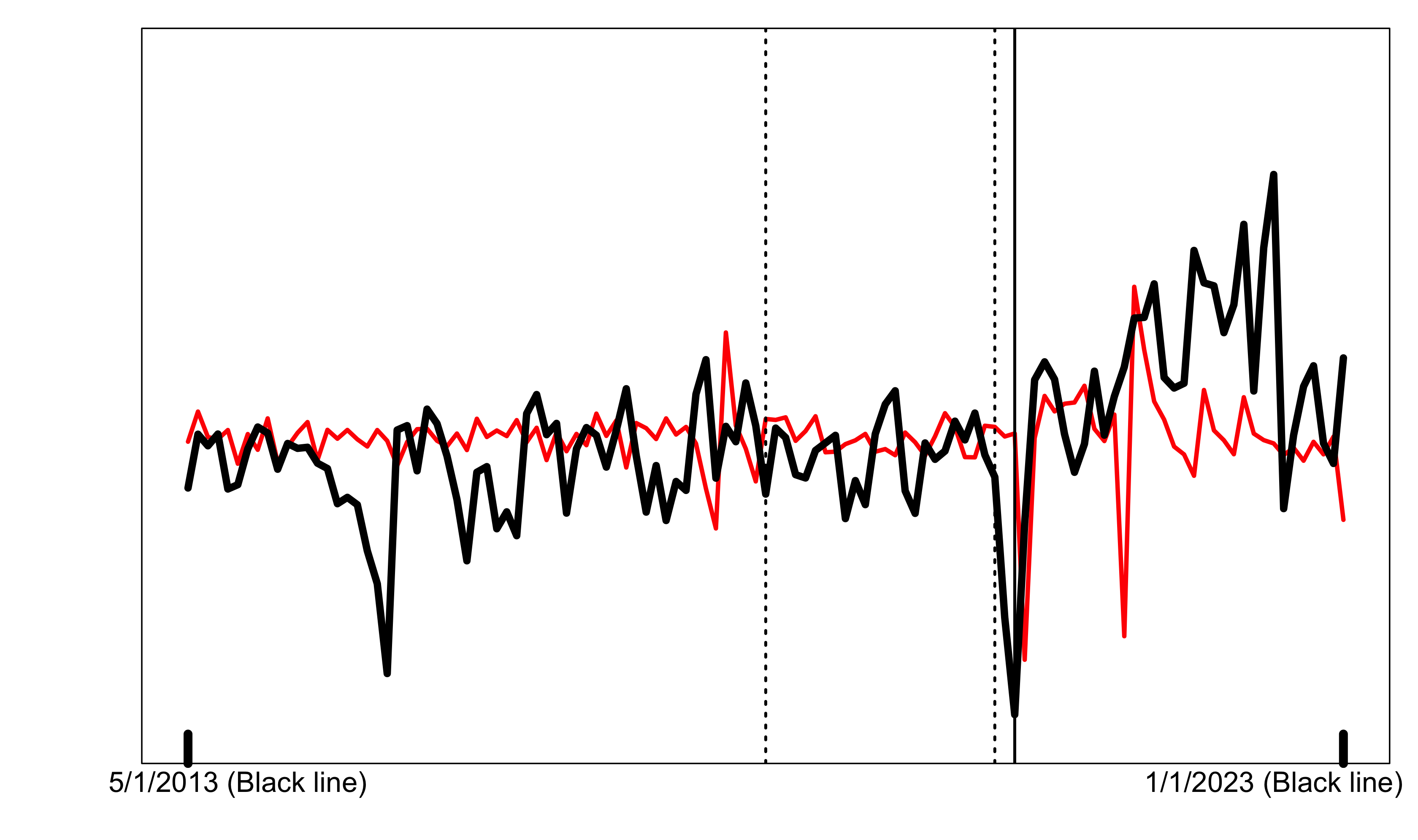}}%
    \caption{The black curves in the three panels are the inflation series at time $t+1$. The red curve in panel (a) indicates the U.S./U.K. exchange rate at time $t$, the red curve in panel (b) is the CPI for durable goods at time $t$, and the red curve in panel (c) is the industrial production index for non-durable goods materials at time $t$. Each pair of two vertical dotted lines indicate an identified significant two-year rolling window, and the vertical solid line is located at April 2020. All curves here are standardized and adjusted for visual comparison, and hence the values of these time series are not reported on the $y$-axis.}
    \label{fig:relations}
\end{figure}

We utilize the stable FACT inference \eqref{reproducible.2} for selecting important covariates from among $p=254$ covariates ($127$ FRED-MD economic variables along with their respective first-order lags) for each rolling window ($94$ windows in total from May 2013 to January 2023, each with $n=24$ observations). We employ the same transformation functions as detailed in Section~\ref{Sec5.4}, with parameters set at $B=1000$ and FDR level $\alpha = 0.2$. Figure~\ref{fig:relations} depicts three significant variables in five rolling windows (no significant discoveries in any other windows), which we specify as follows: 1) U.S./U.K. exchange rate (EXUSUKx) in three rolling windows of January 2018 to December 2019,  April 2018 to March 2020, and October 2018 to September 2020; 2) CPI for durable goods (CUUR0000SAD) from the period of January 2018 to December 2019; and 3) the industrial production index for non-durable goods materials (IPNMAT) from the period of January 2018 to December 2019. The FRED-MD codes of the economic variables are given in the parentheses. Observe that many economic variables exhibit similar behaviors; some of them are  defined similarly and thus highly correlated. Thus, our empirical findings should be carefully interpreted  while keeping the economic intuition and common sense in mind.

In panel a) of Figure~\ref{fig:relations} where EXUSUKx is plotted, the stable FACT inference identifies three significant rolling windows. Each pair of two vertical dotted lines indicate an identified significant two-year rolling window, where different windows are differentiated by thickness of the pair of dotted lines. These discoveries are intuitive since the co-movement 
 of EXUSUKx and the one-month-ahead inflation is visually apparent around 2020. In particular, there are obvious drops for both series. Since the response is the one-month-ahead inflation, the EXUSUKx drops one month earlier in March 2020 than the inflation, making EXUSUKx a one month earlier indicator for the inflation drop. Moreover, our findings suggest that the co-movement between these two series had begun before the COVID-19 pandemic (the vertical solid line at April 2020 in Figure~\ref{fig:relations} indicates the beginning of COVID-19 pandemic), and had persisted over the outbreak of the COVID-19 pandemic.

On the other hand, the other two significant variables, CUUR0000SAD and IPNMAT, are types of consumer price index and industrial production index, respectively. These findings are intuitive since these types of price indices are known to be closely related to the inflation. Interestingly, our empirical analysis suggests that there may be certain nonlinear relations between these variables and the inflation; that is, the e-value statistic $e_{j}(\tanh{(x)}, (\tanh{(x)})^2)$, which measures the nonlinear associations, of the stable FACT for these two variables are quite large. In contrast, our analysis shows that the association between the inflation and EXUSUKx may be mainly linear. Such an empirical result is supported by the clear co-movement between the inflation and EXUSUKx displayed in Figure~\ref{fig:relations}, and the fact that CUUR0000SAD and IPNMAT indeed do not have easily discernible (to the naked eyes) co-movement with the one-month-ahead inflation as shown in Figure~\ref{fig:relations}. 

The above results illustrate how the stable FACT inference can facilitate identifying some important covariates, under the limitation of using only very few observations at each two-year rolling window. Such an advantage can make FACT a powerful and reliable inference tool for different data science applications.

\bibliographystyle{chicago}
\bibliography{references}

	
\newpage
\appendix
\setcounter{page}{1}
\setcounter{section}{0}
\renewcommand{\theequation}{A.\arabic{equation}}
\renewcommand{\thesubsection}{A.\arabic{subsection}}
\setcounter{equation}{0}
	
\begin{center}{\bf \Large Supplementary Material to ``FACT: High-Dimensional Random Forests Inference''}
		
\bigskip
		
Chien-Ming Chi, Yingying Fan and Jinchi Lv
\end{center}
	
\noindent This Supplementary Material contains the proofs of Theorems \ref{theorem3}--\ref{theorem7}, Propositions \ref{prop1}--\ref{prop5}, and some technical lemmas. Hereafter, we denote by $\Phi(\cdot)$ and $\Phi^{-1}(\cdot)$ the cumulative distribution function of the standard Gaussian distribution and its inverse function, respectively. All the notation is the same as defined in the main body of the paper. Throughout the proofs, we use the generic positive constants such as $C$ and $K$ whose values may change from line to line; unless specified otherwise, these constants are independent of the sample size $n$.

	\renewcommand{\thesubsection}{A.\arabic{subsection}}
	
	\section{Proofs of Theorems \ref{theorem3}--\ref{theorem7}} \label{Sec.newA}

	\subsection{Proof of Theorem~\ref{theorem3}} \label{Sec.newA.3}

    Since $K$ and $L$ in Algorithms~\ref{Algorithm1}--\ref{Algorithm2} are constants, we can without loss of generality consider the case when $K=L=1$ (recall that the average of a set of e-values is an e-value). The notation can thus be simplified. In particular, the superscripts and subscripts of the transformation functions and statistics are dropped. To ease the reading, we state the simplified notation here for the proof of Theorem~\ref{theorem3}. Let us define 
\begin{equation}\label{E5.new}
\textnormal{F}_{j} = \frac{\sum_{i\in H_{1}}d_{i}}{ \sqrt{\sum_{i\in H_{1}} (d_{i} - n^{-1}\sum_{i\in H_{1}}d_{i} )^2 } },
\end{equation}
where $d_{i} =  [g_{1}(Y_{i}) - \widehat{Y}(\boldsymbol{X}_{-ij} )]  [g_{2}(X_{ij}) - \widehat{X}(\boldsymbol{X}_{-ij} ) )]$ and $H_{1} = \{1, \dots, n\}$. The random forests model $\widehat{Y}: \mathbb{R}^{p-1} \longmapsto\mathbb{R}$ is obtained by regressing $\{g_{1}(V_{i})\}_{i=1}^{n}$ on $\{\boldsymbol{U}_{-ij}\}_{i=1}^{n}$, while $\widehat{X}: \mathbb{R}^{p-1} \longmapsto\mathbb{R}$ is obtained by regressing $\{g_{2}(U_{ij})\}_{i=1}^{n}$ on $\{\boldsymbol{U}_{-ij}\}_{i=1}^{n}$. 
The forms for the p-value and e-value (see \eqref{E5.3} for details) of $\textnormal{F}_{j}$ are simplified as 
$$P_{j} = 2\Phi(-|\textnormal{F}_{j}| )$$
and
$$ e_{j}(g_{1}, g_{2}) = (P_{j} \vee \epsilon)^{-\frac{1}{2}} - 1,$$ 
respectively, where $0\le\epsilon \le 1$ is as given in Theorem~\ref{theorem3}. 

Furthermore, since $L=1$, we denote by 
$$ \textnormal{FACT}_{j} = e_{j}(g_{1}, g_{2}),$$ 
which is the e-value statistic given in Algorithm~\ref{Algorithm2}.
With such notation, our main goal is to establish that
 \begin{equation}\label{asym.2}
		\mathbb{P}(|\textnormal{F}_{j}| > t ) \le 2\Phi(-t) + Q(t, n, B_{1}, B_{2})
	\end{equation}
    and 
 \begin{equation}
     \label{asym.3}
 \mathbb{E} (\textnormal{FACT}_{j}) \le 1,
 \end{equation}
where
\begin{equation*}
    \begin{split}
        Q(t, n, B_{1}, B_{2}) & \coloneqq \frac{16c}{5\sqrt{\varsigma_{2}}\varsigma_{1}}+ C(n^{-1/4} + B_{1}^{1/4}+ B_{2}^{1/4}) + (-\log{(B_{1}B_{2})})^{-1},\\
        c & = tn^{-1/4}\log{n} + (2t + 1)( 2B_{1}^{1/4} + B_{2}^{1/4}) + \sqrt{nB_{1}B_{2}} (-\log{(B_{1}B_{2})}).
    \end{split}
\end{equation*}
 
	We begin with proving \eqref{asym.2}. Let us decompose test statistic $\textnormal{F}_{j}$  as 
	\begin{equation}\label{thm8.1}
		\begin{split}
			\textnormal{F}_{j} & = \frac{1}{\widehat{\sigma}_{j}\sqrt{n}} \sum_{i=1}^{n}\Big[ (g_{1}(Y_{i}) - \mathbb{E}(g_{1}(Y_{i}) | \boldsymbol{X}_{-ij} ) ) (g_{2}(X_{ij}) - \mathbb{E}(g_{2}(X_{ij} ) | \boldsymbol{X}_{-ij}) ) \\
			& \qquad + (\mathbb{E}(g_{1}(Y_{i}) | \boldsymbol{X}_{-ij} )  - \widehat{Y}(\boldsymbol{X}_{-ij}) ) (g_{2}(X_{ij}) - \mathbb{E}(g_{2}(X_{ij}) | \boldsymbol{X}_{-ij}) )  \\
			&\qquad+ (g_{1}(Y_{i}) - \mathbb{E}(g_{1}(Y_{i}) | \boldsymbol{X}_{-ij} ) ) (\mathbb{E}(g_{2}(X_{ij}) | \boldsymbol{X}_{-ij})  -   \widehat{X} (\boldsymbol{X}_{-ij}) ) \\
			& \qquad+ \underbrace{\Big( (\mathbb{E}(g_{1}(Y_{i}) | \boldsymbol{X}_{-ij} )  - \widehat{Y}(\boldsymbol{X}_{-ij})) (\mathbb{E}(g_{2}(X_{ij}) | \boldsymbol{X}_{-ij})  -  \widehat{X} (\boldsymbol{X}_{-ij})) - \mu \Big)}_{A_{4i}}  + \mu\Big]\\
			& \eqqcolon \frac{1}{\widehat{\sigma}_{j}\sqrt{n}} \sum_{i=1}^{n} (A_{1i} + A_{2i} + A_{3i} + A_{4i} + \mu),
		\end{split}
	\end{equation}
	where 
 \begin{equation*}
     \begin{split}
         \mu & = \mathbb{E}\big\{ \big[\mathbb{E}(g_{1}(Y) | \boldsymbol{X}_{-j} )  - \widehat{Y}(\boldsymbol{X}_{-j}) \big]\big[\mathbb{E}(g_{2}(X_{j}) | \boldsymbol{X}_{-j})  -  \widehat{X} (\boldsymbol{X}_{-j})\big]\big| \mathcal{X}_{0}\big\},\\
         \widehat{\sigma}_{j}^2 & = n^{-1}\sum_{i=1}^n (d_{i} - n^{-1}\sum_{i=1}^n d_{i} )^2  = n^{-1}\sum_{i=1}^n \left(\sum_{k=1}^{4}A_{ki} - n^{-1}\sum_{i=1}^n \sum_{k=1}^{4}A_{ki} \right)^2,
     \end{split}
 \end{equation*}
 and $d_{i}$'s are as given in \eqref{E5.new}. Observe that random variables $A_{1i},A_{2i},A_{3i}$, $A_{4i}$ given in \eqref{thm8.1} above all have zero mean conditional on $\mathcal{X}_{0}$ by construction. 
 
 For a null feature $X_{j}$, it holds that
	\begin{equation*}
	    \begin{split}
	    \sigma_{j}^{2} &\coloneqq \textnormal{Var}\big\{\big[g_{1}(Y)-\mathbb{E}(g_{1}(Y) | \boldsymbol{X}_{-j}) \big] \big[g_{2}(X_{j}) - \mathbb{E}(g_{2}(X_{j}) | \boldsymbol{X}_{-j}) \big] \big\}\\
	    & = \mathbb{E}\big\{\big[g_{1}(Y)-\mathbb{E}(g_{1}(Y) | \boldsymbol{X}_{-j}) \big]\big[g_{2}(X_{j}) - \mathbb{E}(g_{2}(X_{j}) | \boldsymbol{X}_{-j}) \big] \big\}^{2},
	    \end{split}
	\end{equation*}
	which is bounded from above and away from zero due to Lemma~\ref{lower.var.bounds} in Section~\ref{Sec.newC.32} and Condition~\ref{A3}. In view of \eqref{thm8.1}, some simple calculations lead to 
	{\small\begin{equation}\label{thm8.2}
		\begin{split}
			&\Big\{|\textnormal{F}_{j}| > t \Big\} \\
			& \subset \Big\{ \Big|\frac{\sum_{i=1}^{n} A_{1i} }{\sqrt{n}}\Big| > \sigma_{j}t-c+c- |\widehat{\sigma}_{j} - \sigma_{j}|t -\sqrt{n}|\mu|  - n^{-1/2}\Big|\sum_{i=1}^{n} \sum_{l=2,3,4}A_{li}\Big|  \Big\},
		\end{split}
	\end{equation}}
	where $c$ has been defined in \eqref{asym.2}.

	We next build the upper bounds on terms $|\widehat{\sigma}_{j} - \sigma_{j}|$, $\sqrt{n}|\mu|$, and  $n^{-1/2}|\sum_{i=1}^{n} \sum_{l=2,3,4}A_{li}|$ on the right-hand side (RHS) of  \eqref{thm8.2} above. To this end, let us define a number of events 
	\begin{align*}
		& E_{1}^{c} \coloneqq \{ |n^{-1/2}\sum_{i=1}^{n}A_{1i}|\ge n^{1/4}\}, && E_{2}^{c}\coloneqq \{|n^{-1/2}\sum_{i=1}^{n}A_{2i}| \ge B_{1}^{1/4}\},\\
		& E_{3}^{c}\coloneqq \{|n^{-1/2} \sum_{i=1}^{n} A_{3i}|\ge B_{2}^{1/4}\}, \ && E_{4}^{c}\coloneqq \{|n^{-1/2}\sum_{i=1}^{n}A_{4i}| \ge B_{1}^{1/4}\}, \\
		& E_{5}^{c}\coloneqq \{|n^{-1}\sum_{i=1}^{n}A_{2i}^{2}| \ge B_{1}^{1/2}\}, \ &&
		E_{6}^{c}\coloneqq \{|n^{-1}\sum_{i=1}^{n}A_{3i}^{2}| \ge B_{2}^{1/2}\},\\
		& E_{7}^{c} \coloneqq \{|n^{-1}\sum_{i=1}^{n}A_{4i}^{2}| \ge B_{1}^{1/2}\},  && E_{8}^{c} \coloneqq \{  |n^{-1/2} \sum_{i=1}^{n} (A_{1i}^{2} - \sigma_{j}^{2})| \ge n^{1/4}\},\\
		& E_{9}^{c}\coloneqq \{|\mu|\ge \sqrt{B_{1}B_{2}} (-\log{(B_{1}B_{2})})\}
	\end{align*}
	without specifying the dependence on sample size $n$ and convergence rates $B_{1}$ and $B_{2}$. 
 
To deal with term $|\widehat{\sigma}_{j} - \sigma_{j}|$, we construct the upper bounds below. 
First, for any $a, \varepsilon>0$ and $a^{2} \ge \varepsilon$, it holds that 
	\begin{equation}
		\begin{split}
			\label{claim11}
			\sqrt{a^{2} - \varepsilon} \ge \sqrt{(a- \frac{\varepsilon}{a})^{2}} & = a- \frac{\varepsilon}{a},\\
			\sqrt{a^{2} + \varepsilon} \le \sqrt{(a+ \frac{\varepsilon}{a})^{2}} & = a+ \frac{\varepsilon}{a}.
		\end{split}
	\end{equation}
Next, using Minkowski's inequality, we can deduce that 
	{\footnotesize
		\begin{equation}
			\begin{split}
				\label{claim12}
				&\big|\widehat{\sigma}_{j}\big| 
				\ge \sqrt{n^{-1}\sum_{i=1}^{n} A_{1i}^{2}} - \sqrt{n^{-1}\sum_{i=1}^{n} \left(-A_{2i} - A_{3i} - A_{4i} + n^{-1}\sum_{i=1}^{n}(A_{1i} + A_{2i} + A_{3i} + A_{4i}) \right)^{2}}\\
				& \ge \sqrt{n^{-1}\sum_{i=1}^{n} A_{1i}^{2}} - \sum_{l=2}^{4}\sqrt{n^{-1}\sum_{i=1}^{n}A_{li}^{2}} - \sqrt{n^{-1}\sum_{i}\left(n^{-1}\sum_{i=1}^{n}(A_{1i} + A_{2i} + A_{3i} + A_{4i})\right)^{2} }\\
				& \ge \sqrt{\max \left\{ \sigma_{j}^{2} - \left| n^{-1}\sum_{i=1}^{n} (A_{1i}^{2} - \sigma_{j}^{2}) \right|, 0\right\} } - \sum_{l=2}^{4}\sqrt{n^{-1}\sum_{i=1}^{n}A_{li}^{2}} - \Big|n^{-1}\sum_{i=1}^{n}(A_{1i} + A_{2i} + A_{3i} + A_{4i})\Big|.
			\end{split}
	\end{equation}}
	Similarly, we can show that 
	\begin{equation}
		\begin{split}
			\label{claim13}
			|\widehat{\sigma}_{j}| & \le \sqrt{\sigma_{j}^{2} + \left| n^{-1}\sum_{i=1}^{n} (A_{1i}^{2} - \sigma_{j}^{2}) \right| } + \sum_{l=2}^{4}\sqrt{n^{-1}\sum_{i=1}^{n}A_{li}^{2}} \\
			&\quad+ \Big|n^{-1}\sum_{i=1}^{n}(A_{1i} + A_{2i} + A_{3i} + A_{4i})\Big|.
		\end{split}
	\end{equation}

 From the definition of $\widehat{\sigma}_{j}^{2}$ and (\ref{claim11})--(\ref{claim13}), it holds that if $\sigma_{j}^{2}\ge \left| n^{-1}\sum_{i=1}^{n} (A_{1i}^{2} - \sigma_{j}^{2}) \right|$,
	\begin{equation}
		\label{thm8.3}
		\begin{split}
			& |\widehat{\sigma}_{j} - \sigma_{j}| \le \left|\frac{1}{n\sigma_{j}}\sum_{i=1}^{n} (A_{1i}^{2} - \sigma_{j}^{2}) \right| \\
			& \quad+ \sum_{l=2}^{4}\sqrt{n^{-1}\sum_{i=1}^{n}A_{li}^{2}} + |n^{-1}\sum_{i=1}^{n}(A_{1i} + A_{2i} + A_{3i} + A_{4i})|.
		\end{split}
	\end{equation}
	In light of \eqref{thm8.3}, it holds on event $\cap_{l=1}^{8}E_{l}$ that for all large $n$, \[ |\widehat{\sigma}_{j} - \sigma_{j}|\le n^{-1/4}\log{ n} + 4B_{1}^{1/4} + 2B_{2}^{1/4} . \] 
	On event $\cap_{l=2}^{4}E_{l}$, we can show that for all large $n$, 
	\[ n^{-1/2}|\sum_{i=1}^{n} \sum_{l=2,3,4}A_{li}|\le 2B_{1}^{1/4} + B_{2}^{1/4}. \] 
	Moreover, on event $E_{9}$, it holds that 
	$$ \sqrt{n}|\mu|\le \sqrt{nB_{1}B_{2}} (-\log{(B_{1}B_{2})}).$$

	From these upper bounds above, the definition of $c$, and \eqref{thm8.2}, it holds on event $\cap_{l=1}^{9} E_{l}$ that for all large $n$, each $t>0$, and each $B_{1}, B_{2}>0$,
	\begin{equation}\label{thm8.4}
		\Big\{|\textnormal{F}_{j}| > t  \Big\} \subset \Big\{ |\frac{\sum_{i=1}^{n} A_{1i} }{\sqrt{n}}| > \sigma_{j}t - c \Big\},
	\end{equation}
	where we recall that 
	$$c = tn^{-1/4}\log{n} + (2t + 1)( 2B_{1}^{1/4} + B_{2}^{1/4}) + \sqrt{nB_{1}B_{2}} (-\log{(B_{1}B_{2})}).$$
We can further deduce that for some $C>0$, all large $n$, each $t > 0$, and each $B_{1}, B_{2}>0$, 
	\begin{equation}
 \begin{split}
		\label{thm8.5}
		& \mathbb{P}(|\textnormal{F}_{j}| > t ) \\ & \le \sum_{l=1}^{9}\mathbb{P} (E_{l}^{c}) +  \mathbb{P}\big( |\frac{\sum_{i=1}^{n} A_{1i} }{\sqrt{n}}| > \sigma_{j}t - c  \big) \\
  &\le\sum_{l=1}^{9}\mathbb{P} (E_{l}^{c})+  \Big( 1 - \mathbb{P}\big( \frac{\sum_{i=1}^{n} A_{1i} }{\sqrt{n}} \le \sigma_{j}t - c  \big)\Big) + \mathbb{P}\big( \frac{\sum_{i=1}^{n} A_{1i} }{\sqrt{n}}  < -\sigma_{j}t + c \big) \\
  & \le \sum_{l=1}^{9}\mathbb{P} (E_{l}^{c}) \\
  & \qquad + (1 - \Phi(t)) + |\Phi(t) - \Phi(t- \frac{c}{\sigma_{j}})| + \Big|\Phi(t- \frac{c}{\sigma_{j}}) - \mathbb{P}\big( \frac{\sum_{i=1}^{n} A_{1i} }{\sqrt{n}} \le \sigma_{j}t - c  \big)\Big| \\
  & \qquad + \Phi(-t) + |\Phi(-t + \frac{c}{\sigma_{j}}) - \Phi(-t) | + \Big|\mathbb{P}\big( \frac{\sum_{i=1}^{n} A_{1i} }{\sqrt{n}}  < -\sigma_{j}t + c \big) - \Phi(-t + \frac{c}{\sigma_{j}}) \Big|  \\
&\le 2\Phi(-t) +  |\Phi(t) - \Phi(t- \frac{c}{\sigma_{j}})| \\
& \qquad + |\Phi(-t + \frac{c}{\sigma_{j}}) - \Phi(-t) | + Cn^{-1/3} + \sum_{l=1}^{9}\mathbb{P} (E_{l}^{c}) \\
 & \le 2\Phi(-t) + \frac{16c}{5\sqrt{\varsigma_{2}}\varsigma_{1}} + Cn^{-1/3} + \sum_{l=1}^{9}\mathbb{P} (E_{l}^{c}),
	\end{split} 
 \end{equation} 
where the first inequality above is from \eqref{thm8.4}, the fourth inequality above follows from Lemma~\ref{claim2} in Section~\ref{Sec.newC.2} and the assumption of $\mathbb{E} [g_{1}(Y)]^{4} < D_{2}$ (Condition~\ref{A3}), and the last inequality above is due to the property of the Gaussian distribution that for each  $\sigma>0$ and $t, x\in \mathbb{R}$, 
	\begin{equation}
	    \begin{split}
	        \label{gaussian.difference.1}
	        \left|\Phi(t + \frac{x}{\sigma}) - \Phi(t)\right|\le 0.4 \times \frac{|x|}{\sigma}
	    \end{split}
	\end{equation}
	and that 
\begin{equation}
	    \begin{split}
	        \label{gaussian.difference.2}
	        0.4 \frac{c}{\sigma_{j}} \le \frac{8c}{5\sqrt{\varsigma_{2}}\varsigma_{1}},
	    \end{split} 
	\end{equation}
 which follows from Condition~\ref{A3} and Lemma~\ref{lower.var.bounds}.

	It remains to upper bound the probabilities $\mathbb{P} (E_{1}^{c}), \dots, \mathbb{P} (E_{9}^{c})$ that appear in \eqref{thm8.5} above. Let us begin with the bound for term $\mathbb{P}(E_{3}^{c})$. Note that we have $\mathbb{E}(A_{3i} | \mathcal{X}_{0}) = 0$. Then using the Markov inequality, the Burkholder--Davis--Gundy inequality \citep{burkholder1972integral}, and Jensen's inequality, we can show that there exists some $C >0$ such that for all large $n$ and each $B_{2} > 0$,
	{\small\begin{equation}\begin{split}
			\label{thm8.7}
			&\mathbb{P}(E_{3}^{c}) \le B_{2}^{-1/4}\mathbb{E}|n^{-1/2}\sum_{i=1}^{n}A_{3i}|  \le B_{2}^{-1/4}C\Big(n^{-1}\sum_{i=1}^{n}\mathbb{E}(A_{3i})^{2}\Big)^{1/2}.
        \end{split}\end{equation}}%
        Furthermore, it holds that 
   {\small\begin{equation}\begin{split}
   \label{thm8.7b}
		& \textnormal{RHS of \eqref{thm8.7}}	
  \\ &  = B_{2}^{-1/4}C\Big\{\mathbb{E} \Big\{ \big[g_{1}(Y) - \mathbb{E}(g_{1}(Y) | \boldsymbol{X}_{-j} ) \big]^2 \big[\mathbb{E}(g_{2}(X_{j}) | \boldsymbol{X}_{-j})  -   \widehat{X} (\boldsymbol{X}_{-j}) \big]^2   \Big\}\Big\}^{1/2}\\
			& = B_{2}^{-1/4}C\Big\{\mathbb{E} \Big\{ \mathbb{E}\big[ (g_{1}(Y) - \mathbb{E}(g_{1}(Y) | \boldsymbol{X}_{-j} ) )^2 \big| \mathcal{X}_{0}, \boldsymbol{X}_{-j}\big] \big[\mathbb{E}(g_{2}(X_{j}) | \boldsymbol{X}_{-j}) - \widehat{X} (\boldsymbol{X}_{-j}) \big]^2   \Big\} \Big\}^{1/2}\\
			& \le B_{2}^{-1/4}D^{1/2}C\Big\{\mathbb{E}   \big[\mathbb{E}(g_{2}(X_{j}) | \boldsymbol{X}_{-j}) - \widehat{X} (\boldsymbol{X}_{-j}) \big]^2    \Big\}^{1/2}\\
			&\le B_{2}^{1/4}D^{1/2}C,
	\end{split}\end{equation}}%
	where the first equality above is due to the fact that $(\boldsymbol{X}, g_{1}(Y))$ and $(\boldsymbol{X}_{i}, g_{1}(Y_{i}))$ have the same distribution for each $i$, the second equality above is because $\mathbb{E}(g_{2}(X_{j}) | \boldsymbol{X}_{-j})$ and $\widehat{X} (\boldsymbol{X}_{-j})$ are $\sigma(\mathcal{X}_{0}, \boldsymbol{X}_{-j})$-measurable, the first inequality above is entailed by the assumptions that $\mathbb{E}\big\{(g_{1}(Y) - \mathbb{E}(g_{1}(Y)| \boldsymbol{X}_{-j}) )^{2}| \boldsymbol{X}_{-j}\big\}$ $= \textnormal{Var}(g_{1}(Y)|\boldsymbol{X}_{-j})\le D$ (see Condition~\ref{A3}) and that $\mathcal{X}_{0}$ is an independent sample, and the last inequality above utilizes Condition~\ref{A7}. 
	
	The arguments for the rest of the upper bounds are similar to those for \eqref{thm8.7}--\eqref{thm8.7b}. Hence, we will omit the technical details here for simplicity and stress only which consistency condition among Condition~\ref{A1}, Condition~\ref{A7}, and \eqref{high.moment.bound.1} below will be needed for each upper bound. Because random forests makes predictions via the conditional sample averages and we have assumed $0\le g_{2}(X_{j})\le 1$, it follows that $0\le \widehat{X}( \boldsymbol{X}_{-j}) \le 1$. In view of $0\le \widehat{X}( \boldsymbol{X}_{-j}) \le 1$, Condition~\ref{A1}, and the assumption that $0\le g_{2}(X_{j})\le 1$ in Condition~\ref{A3}, we can deduce that 
	\begin{equation}
	    \begin{split}\label{high.moment.bound.1}
	    &\mathbb{E}\big\{ [\mathbb{E}(g_{1}(Y) |\boldsymbol{X}_{-j}) - \widehat{Y}(\boldsymbol{X}_{-j}) ] [\mathbb{E}(g_{2}(X_{j}) |\boldsymbol{X}_{-j}) - \widehat{X}( \boldsymbol{X}_{-j}) ]  - \mu\big\}^{2} \\
	        &\le \mathbb{E}\big\{ [\mathbb{E}(g_{1}(Y) |\boldsymbol{X}_{-j}) - \widehat{Y}(\boldsymbol{X}_{-j}) ] [\mathbb{E}(g_{2}(X_{j}) |\boldsymbol{X}_{-j}) - \widehat{X}( \boldsymbol{X}_{-j}) ]  \big\}^{2} \\
	        &\le \mathbb{E} [\mathbb{E}(g_{1}(Y) |\boldsymbol{X}_{-j}) - \widehat{Y}(\boldsymbol{X}_{-j}) ]^2 \\
	        &\le B_{1}.
	    \end{split}
	\end{equation}

	It follows from \eqref{high.moment.bound.1} and other regularity conditions that there exists some $K_{0} >0$ such that for all large $n$ and each $B_{1} > 0$,
	\begin{equation}
		\label{thm8.9}
		\mathbb{P}(E_{4}^{c}) 
		\le K_{0}B_{1}^{1/4}.
	\end{equation}
	By Condition~\ref{A1} and other regularity conditions, there exists some $K_{0} >0$ such that for all large $n$ and each $B_{1} > 0$,
	\begin{equation}
		\label{thm8.6}
		\mathbb{P}(E_{2}^{c})  \le
		K_{0}B_{1}^{1/4}.
	\end{equation}
	From Condition~\ref{A7} and other regularity conditions, we can show that for all large $n$ and $B_{2}>0$,
	\begin{equation}
		\label{thm8.11}
		\mathbb{P}(E_{6}^{c}) \le  DB_{2}^{1/2}.
	\end{equation}
	
	Furthermore, by Condition \ref{A1} and other regularity conditions, we can deduce that for all large $n$ and $B_{1}>0$,
	\begin{equation}
		\label{thm8.10}
		\mathbb{P}(E_{5}^{c})  \le B_{1}^{1/2}.
	\end{equation}
	An application of \eqref{high.moment.bound.1} and other regularity conditions yields that for all large $n$ and $B_{1}>0$,
	\begin{equation}
		\label{thm8.12}
		\mathbb{P}(E_{7}^{c})  \le 4B_{1}^{1/2}.
	\end{equation}
	In addition, by the assumptions, we can show that there exists some $C>0$ such that for all large $n$,
	\begin{equation}\begin{split}
			\label{thm8.13}
			\mathbb{P}(E_{1}^{c})& \le Cn^{-1/4},\\
			\mathbb{P}(E_{8}^{c})& \le Cn^{-1/4}.
		\end{split}
	\end{equation}
	With the aid of the Markov inequality, Jensen's inequality, the Cauchy--Schwartz inequality, and Conditions~\ref{A1}--\ref{A7}, it holds that for all $n\ge 1$, each $B_{1}, B_{2}>0$, and each $t>0$,
	\begin{equation}
	    \begin{split}\label{thm8.14}
	         \mathbb{P}(E_{9}^{c}) &\le \frac{ \mathbb{E}\Big|\mathbb{E}\big\{ \big[\mathbb{E}(g_{1}(Y) | \boldsymbol{X}_{-j} )  - \widehat{Y}(\boldsymbol{X}_{-j}) \big]\big[\mathbb{E}(g_{2}(X_{j}) | \boldsymbol{X}_{-j})  -  \widehat{X} (\boldsymbol{X}_{-j})\big] \big| \mathcal{X}_{0}\big\} \Big|}{(-\log{(B_{1}B_{2})})\sqrt{B_{1}B_{2}}}\\
	        & \le (-\log{(B_{1}B_{2})})^{-1}.
	    \end{split}
	\end{equation}
	
Thus, combining the probabilities bounds in \eqref{thm8.5}--\eqref{thm8.14} above and the fact that $n^{-1/3} = o(n^{-1/4})$, we can obtain that there exists some $C>0$ such that for all large $n$, all consistency rates $0<B_{1},B_{2}<1$ (note that $x^{1/2}\le x^{1/4}$ if $0<x<1$), and each positive test threshold level $t$,
    \begin{equation} \label{evalue.3}
		\mathbb{P}(|\textnormal{F}_{j}| > t ) \le 2\Phi(-t) + \frac{16c}{5\sqrt{\varsigma_{2}}\varsigma_{1}}+ C(n^{-1/4} + B_{1}^{1/4}+ B_{2}^{1/4}) + (-\log{(B_{1}B_{2})})^{-1},
    \end{equation}
which finishes the proof for \eqref{asym.2}.

We now proceed with establishing\eqref{asym.3}. The result of \eqref{evalue.3} implies that for each $z \ge 0$,
\begin{equation}
    \begin{split}\label{evalue.1}
        \mathbb{P}(P_{j} \le z ) 
        & = \mathbb{P}\left( 2\Phi(-|\textnormal{F}_{j}| )\le z \right)\\
        & = \mathbb{P}\left( |\textnormal{F}_{j}|   \ge -\Phi^{-1}\left(\frac{z}{2} \right) \right) \\    
        &\le z + Q(-\Phi^{-1}\left(\frac{z}{2}\right), n, B_{1}, B_{2}),
    \end{split}
\end{equation}
where $P_{j}$ is given in \eqref{E5.3}, $Q(t, n, B_{1}, B_{2})$ has been defined in \eqref{asym.2}, and $\Phi(\cdot)$ and $\Phi^{-1}(\cdot)$ denote the cumulative distribution function of the standard Gaussian distribution and its inverse function, respectively.

With sufficiently large $n$ and the choice of  $\epsilon = (\log{n})^{-1}$, a direct calculation shows that 
\begin{equation}
    \begin{split}\label{evalue.2}
        \mathbb{E}(\textnormal{FACT}_{j}) & = \mathbb{E}((P_{j} \vee \epsilon)^{-\frac{1}{2}} - 1 ) \\
        & = \int_{0}^{\infty} \mathbb{P}(P_{j} \vee \epsilon)^{-\frac{1}{2}} \ge z  )dz - 1 \\
        & \le \int_{1}^{\infty} \mathbb{P}(P_{j} \vee \epsilon)^{-\frac{1}{2}} \ge z  )dz\\
        & = \int_{1}^{\epsilon^{-\frac{1}{2}}} \mathbb{P}(P_{j}^{-\frac{1}{2}} \ge z  )dz \\
        &\le \int_{1}^{\epsilon^{-\frac{1}{2}}} z^{-2} + Q\left(-\Phi^{-1}\left(\frac{\epsilon}{2}\right), n, B_{1}, B_{2}\right) dz\\
        & = (-\sqrt{\epsilon} + 1) + (\epsilon^{-\frac{1}{2}} - 1) \times Q\left(-\Phi^{-1}\left(\frac{\epsilon}{2}\right), n, B_{1}, B_{2}\right) \\
        & \le 1- Q\left(-\Phi^{-1}\left(\frac{\epsilon}{2}\right), n, B_{1}, B_{2}\right),
    \end{split}
\end{equation}
where the first equality above holds because $\mathbb{E}(X) = \int_{0}^{\infty}\mathbb{P}(X\ge z) dz$ when $X\ge 0$ almost surely, the first inequality above holds because the probability measure is bounded by one, the second inequality above follows from \eqref{evalue.1} and the definitions of $Q(\cdot)$ and $\Phi^{-1}(\cdot)$, and the last equality above holds because $Q\left(-\Phi^{-1}\left(\frac{\epsilon}{2}\right), n, B_{1}, B_{2}\right)\times \epsilon^{-1} \le 1$ for all large $n$, which is derived from 
the assumption that $(B_{1}+ B_{2})(\log{n})^2\sqrt{n} = o(1)$ and the choice of $\epsilon>0$ such that $\epsilon = (\log{n})^{-1}$. 
Therefore, by \eqref{evalue.2} and the fact that $Q\left(-\Phi^{-1}\left(\frac{\epsilon}{2}\right), n, B_{1}, B_{2}\right)\ge 0$ for all large $n$, we obtain the desired conclusion in  \eqref{asym.3}. This completes the proof of Theorem~\ref{theorem3}.

	\subsection{Proof of Theorem~\ref{theorem7}} \label{Sec.newA.7}

For the reader's convenience, we reiterate the needed notation here for the proof of Theorem~\ref{theorem7}. Let $H_{1}, \dots, H_{K}$ be a partition of the index set $\{1, \dots, n\}$ such that  $H_{k}\cap H_{l} = \emptyset$ and $|\# H_{k} - \# H_{l}|\le 1$ for all $\{k, l\} \subset \{1, \dots,K\}$ with $k\not= l$. For each $k \in \{1,\dots ,K\}$, let us define
\begin{equation}\label{E5.new}
\textnormal{F}_{j, l}^{(k)} = \frac{\sum_{i\in H_{k}}d_{il}}{ \sqrt{\sum_{i\in H_{k}} (d_{il} - (\# H_{k})^{-1}\sum_{i\in H_{k}}d_{il} )^2 } },
\end{equation}
where $d_{il} =  [g_{1l}(Y_{i}) - \widehat{Y}(\boldsymbol{X}_{-ij} )]  [g_{2l}(X_{ij}) - \widehat{X}(\boldsymbol{X}_{-ij} ) ]$. The dependence on $l\in\{1, 2\}$ ($L=2$ here) is specified in \eqref{E5.new}. The random forests model $\widehat{Y}: \mathbb{R}^{p-1} \longmapsto\mathbb{R}$ is obtained by regressing $\{g_{1l}(V_{i})\}_{i=1}^{n}$ on $\{\boldsymbol{U}_{-ij}\}_{i=1}^{n}$, while $\widehat{X}: \mathbb{R}^{p-1} \longmapsto\mathbb{R}$ is obtained by regressing $\{g_{2l}(U_{ij})\}_{i=1}^{n}$ on $\{\boldsymbol{U}_{-ij}\}_{i=1}^{n}$, where $\{V_{i}, \boldsymbol{U}_{i}\}$ is the training sample (also see Algorithm~\ref{Algorithm1}).
In addition, the expressions for the p-value and e-value (see \eqref{E5.3} for details) of $\textnormal{F}_{j, l}^{(k)}$ are given by
 \begin{equation*}
     P_{j, l}^{(k)} \coloneqq 2\Phi(-|\textnormal{F}_{j, l}^{(k)}| ), \qquad e_{j,l}^{(k)} \coloneqq (P_{j, l}^{(k)} \vee \epsilon)^{-\frac{1}{2}} - 1, \qquad e_{j}(g_{1l}, g_{2l}) \coloneqq K^{-1}\sum_{k=1}^K {e}_{j, l}^{(k)},
 \end{equation*}
respectively, where $\epsilon =0$ is assumed by Theorem~\ref{theorem7}. Hence, it follows that 
$$ \textnormal{FACT}_{j} = \frac{1}{L} \sum_{l=1}^L e_{j}(g_{1l}, g_{2l}) = \frac{1}{L} \sum_{l=1}^L \frac{1}{K}\sum_{k=1}^K  {e}_{j, l}^{(k)} ,$$ 
which is the e-value statistic given in Algorithm~\ref{Algorithm2}.

	In light of the assumption $\sum_{s=1}^2|\kappa_{s}| > 0$, let us assume without loss of generality that $|\kappa_{1}| > 0$. Then, with $\textnormal{FACT}_{j}$ given above, $L=2$, and tuning parameter $\epsilon=0$, we can deduce that 
 \begin{equation}
 \begin{split}
     \label{power.7.new}
     \mathbb{E}(\textnormal{FACT}_{j}) & \ge \frac{1}{KL} \times \mathbb{E}\left(\frac{1}{\sqrt{2\Phi(-|\textnormal{F}_{j,1}^{(1)}|)} } - 1\right) \\
     & \ge 
    \frac{1}{2K} \times \sup_{z\ge 0} \left\{ (z-1) \times 
     \left[1 - \mathbb{P}\left(\frac{1}{\sqrt{2\Phi(-|\textnormal{F}_{j,1}^{(1)}|)} } \le z\right)\right] \right\}\\
     & = \frac{1}{2K} \times \sup_{z\ge 0} \left\{(z-1) \times 
     \left[1 - \mathbb{P}\left(|\textnormal{F}_{j,1 }^{(1)}| \le -\Phi^{-1}(\frac{z^{-2}}{2})\right) \right]\right\},
     \end{split}
 \end{equation}
	where $\textnormal{F}_{j, 1}^{(1)}$ is defined in \eqref{E5.new} with transformations $ g_{11}(x) =g_{21}(x) = x$. To further simplify the notation, let us assume that $H_{1} = \{1, \dots, n_{1}\}$ with $n_{1}$ the smallest integer such that $n_{1}\ge \frac{n}{K}$, and we 
 $$\textnormal{denote }  F_{j, 1}^{(1)}  \textnormal{ above as } F_{j}.$$

We next analyze the RHS of \eqref{power.7.new} above. It holds that for each $t\ge 0$,
	\begin{equation}\label{thm7.1}
		\begin{split}
			& \mathbb{P}\left(  | \textnormal{F}_{j}| \le t  \right) \\
			& \le \mathbb{P}\left( \left| \textnormal{F}_{j} -\sqrt{n_{1}}\frac{\kappa_{1}}{\widehat{\sigma}_{j}} + \sqrt{n_{1}}\frac{\kappa_{1}}{\widehat{\sigma}_{j}}\right| \le t  \right) \\
			& \le  \mathbb{P}\left(  \sqrt{n_{1}}\frac{\left|\kappa_{1}\right| }{\widehat{\sigma}_{j}}  \le t + \left| \textnormal{F}_{j} -\sqrt{n_{1}}\frac{\kappa_{1}}{\widehat{\sigma}_{j}} \right| \right),
		\end{split}
	\end{equation}
		where $\widehat{\sigma}_{j}^2  = n_{1}^{-1}\sum_{i\in H_{1}} (d_{i1} - n^{-1}\sum_{i\in H_{1}} d_{i1} )^2$.
	We further bound the RHS of \eqref{thm7.1} above, where $g_{11}(x) = g_{21}(x) = x$ is used throughout the arguments below. By the definition of $\textnormal{F}_{j}$ above, let us define terms $A_{5i}, \dots, A_{8i}$ as 
	\begin{equation}\label{thm7.3}
		\begin{split}
			& \textnormal{F}_{j} -\sqrt{n_{1}}\frac{\kappa_{1}}{\widehat{\sigma}_{j}} \\
			& = \frac{1}{\widehat{\sigma}_{j}\sqrt{n_{1}}} \sum_{i=1}^{n_{1}}\Big[ \Big((Y_{i} - \mathbb{E}(Y_{i} | \boldsymbol{X}_{-ij} ) ) (X_{ij} - \mathbb{E}(X_{ij}  | \boldsymbol{X}_{-ij}) - \kappa_{1} \Big) \\
			& \quad+ (\mathbb{E}(Y_{i} | \boldsymbol{X}_{-ij} )  - \widehat{Y}(\boldsymbol{X}_{-ij}) ) (X_{ij} - \mathbb{E}(X_{ij} | \boldsymbol{X}_{-ij}) ) \\
			&\quad+ (Y_{i} - \mathbb{E}(Y_{i} | \boldsymbol{X}_{-ij} ) ) (\mathbb{E}(X_{ij} | \boldsymbol{X}_{-ij}) - \widehat{X}(\boldsymbol{X}_{-ij}) ) \\
			& \quad+ (\mathbb{E}(Y_{i} | \boldsymbol{X}_{-ij} )  - \widehat{Y}(\boldsymbol{X}_{-ij}) ) (\mathbb{E}(X_{ij} | \boldsymbol{X}_{-ij}) - \widehat{X}(\boldsymbol{X}_{-ij}) ) \Big]\\
			& \eqqcolon \frac{1}{\widehat{\sigma}_{j}\sqrt{n_{1}}} \sum_{i=1}^{n_{1}} (A_{5i} + A_{6i} + A_{7i} + A_{8i}).
		\end{split}
	\end{equation}

	From \eqref{thm7.3} and the Markov inequality, we can show that 
	\begin{equation}
	    \begin{split}
	        & \mathbb{P}\left(  \sqrt{n_{1}}\frac{\left|\kappa_{1}\right| }{\widehat{\sigma}_{j}}  \le t + \left| \textnormal{F}_{j} -\sqrt{n_{1}}\frac{\kappa_{1}}{\widehat{\sigma}_{j}} \right| \right) \\
	        & \le \mathbb{P}\left(  \sqrt{n_{1}} \left|\kappa_{1}\right|   \le t\widehat{\sigma}_{j} + \sum_{k=5}^8 \left|\frac{\sum_{i=1}^{n_{1}} A_{ki} }{\sqrt{n_{1}}} \right|  \right)\\
	        & \le \frac{t\mathbb{E}(\widehat{\sigma}_{j}) + \sum_{k=5}^8 \mathbb{E}\left|\frac{\sum_{i=1}^{n_{1}} A_{ki} }{\sqrt{n_{1}}} \right|}{\sqrt{n_{1}} \left|\kappa_{1}\right|}.
	    \end{split}
	\end{equation}
Recall that a random forests estimate is some average of the training sample. Since $\widehat{X}(\boldsymbol{X}_{-j})$ is a random forests estimate of $\mathbb{E}(X_{j}|\boldsymbol{X}_{-j})$ and $0 \le X_{j}\le 1$ by assumption, it follows that $0\le \widehat{X}(\boldsymbol{X}_{-j}) \le 1$. By this result, the assumptions of i.i.d. observations, Jensen's inequality, and Condition~\ref{A1}, it holds that for each $t>0$, $B_{1}, B_{2}>0$, and all $n_{1}\ge 1$,
	\begin{equation}
	    \begin{split}
	        \mathbb{E}(\widehat{\sigma}_{j}) & \le \sqrt{\mathbb{E}(\widehat{\sigma}_{j})^2}\\
	        & \le \sqrt{\mathbb{E}\big\{\big[Y - \widehat{Y}(\boldsymbol{X}_{-j} ) \big] \big[X_{j} - \widehat{X}(\boldsymbol{X}_{-j}) \big] \big\}^2}\\
	        & \le \sqrt{\mathbb{E}(Y - \widehat{Y}(\boldsymbol{X}_{-j} ) )^2}\\
	        & = \sqrt{\mathbb{E}(Y -\mathbb{E}(Y|\boldsymbol{X}_{-j}) + \mathbb{E}(Y|\boldsymbol{X}_{-j})- \widehat{Y}(\boldsymbol{X}_{-j} ) )^2}\\
	        &\le \sqrt{\textnormal{Var}(Y) + B_{1}}.
	    \end{split}
	\end{equation}

	With the aid of the Burkholder--Davis--Gundy inequality, Jensen's inequality, the assumptions of i.i.d. observations and $0\le X_{j} \le 1$, there exists some constant $C_{2}>0$ such that for each $t>0$, $B_{1}, B_{2}>0$, and all $n_{1}\ge 1$,
	\begin{equation}
	    \begin{split}
	    \mathbb{E}\left|\frac{\sum_{i=1}^{n_{1}} A_{5i} }{\sqrt{n_{1}}} \right| &\le C_{2} \sqrt{\mathbb{E}\Big((Y - \mathbb{E}(Y | \boldsymbol{X}_{-j} ) ) (X_{j} - \mathbb{E}(X_{j}  | \boldsymbol{X}_{-j})) - \kappa_{1} \Big)^2 }\\
	    & \le C_{2}\sqrt{\textnormal{Var}(Y)}.
	    \end{split}
	\end{equation}
	From the Burkholder--Davis--Gundy inequality, Jensen's inequality, the assumptions of i.i.d. observations and $0\le X_{j} \le 1$, and Condition~\ref{A1}, we can show that there exists some constant $C>0$ such that for each $t>0$, $B_{1}>0$, and all $n_{1}\ge 1$,
	\begin{equation}
	    \begin{split}
	    \mathbb{E}\left|\frac{\sum_{i=1}^{n_{1}} A_{6i} }{\sqrt{n_{1}}} \right| &\le C\sqrt{B_{1}}.
	    \end{split} 
	\end{equation}
	
	Furthermore, by the Burkholder--Davis--Gundy inequality, Jensen's inequality, the assumptions of i.i.d. observations and $\textnormal{Var}(Y|\boldsymbol{X}_{-j})\le D$ almost surely (see the second assumption of Theorem~\ref{theorem7}, and recall that the transformation functions $g_{11}(x) = g_{21}(x) = x$ are used here), the assumption that the training sample is an independent sample, and Condition~\ref{A7}, there exists some constant $C>0$ such that for each $t>0$, $B_{1}, B_{2}>0$, and all $n_{1}\ge 1$,
	\begin{equation}
	    \begin{split}
	        \mathbb{E}\left|\frac{\sum_{i=1}^{n_{1}} A_{7i} }{\sqrt{n_{1}}} \right| &\le C\sqrt{\mathbb{E} \big[(Y - \mathbb{E}(Y | \boldsymbol{X}_{-j} ) ) (\mathbb{E}(X_{j} | \boldsymbol{X}_{-j}) - \widehat{X}(\boldsymbol{X}_{-j}) ) \big]^2 }\\
	        &\le C\sqrt{D\mathbb{E}  (\mathbb{E}(X_{j} | \boldsymbol{X}_{-j}) - \widehat{X}(\boldsymbol{X}_{-j}) )^2 }\\
	        &\le C\sqrt{DB_{2}} ,
	    \end{split}
	\end{equation}
	where the second inequality above uses an argument similar to that for \eqref{thm8.7}--\eqref{thm8.7b}. By Conditions~\ref{A1} and \ref{A7}, the assumptions of i.i.d. observations, and the Cauchy–Schwartz inequality, it holds that for each $t>0$, $B_{1}, B_{2}>0$, and all $n_{1}\ge 1$,
	\begin{equation}
	    \begin{split}\label{thm7.4}
	        \mathbb{E}\left|\frac{\sum_{i=1}^{n_{1}} A_{8i} }{\sqrt{n_{1}}} \right| &\le \sqrt{n_{1}B_{1}B_{2}}.
	    \end{split}
	\end{equation}
	Thus, in light of \eqref{thm7.1}--\eqref{thm7.4} and the property of the subadditivity inequality, there exists some $C>0$ such that for each $t>0$, $B_{1}, B_{2} >0$, and all $n_{1}\ge 2$,
	\begin{equation}
	    \begin{split}\label{power.1.new}
	         \mathbb{P}\left(  | \textnormal{F}_{j} |\le t  \right) &\le \frac{(C+t)(\textnormal{Var}(Y) + \sqrt{B_{1}} +\sqrt{B_{2}} + \sqrt{n_{1}B_{1}B_{2}}) }{\sqrt{n_{1}} \left|\kappa_{1}\right|}.
	    \end{split}
	\end{equation}

To deal with the RHS of \eqref{power.7.new}, we will need \eqref{power.1.new} above and two results in \eqref{power.5.new}--\eqref{power.6.new} presented below. For each $x\ge 0$, we have 
\begin{equation}
    \begin{split}\label{power.5.new}
         \Phi(-x) \le \frac{1}{\sqrt{2\pi}} \left(\frac{1}{x} \right) \exp(\frac{-x^2}{2}),
    \end{split}
\end{equation}
which is used to conclude that for all large $z$,
\begin{equation}
    \begin{split}\label{power.6.new}
2\sqrt{\log{z}} \ge - \Phi^{-1}\left(\frac{1}{\sqrt{8\pi \log{z}}}  \exp{\left(\frac{- 4\log{z}}{2} \right)} \right) \ge -\Phi^{-1}(\frac{z^{-2}}{2}).
\end{split}
\end{equation}
From \eqref{power.1.new} and \eqref{power.6.new}, the assumption that $B_{1}B_{2}n +B_{1} + B_{2}\le 1$, and Condition~\ref{A3}, there exists some $C>0$ such that for each $B_{1}, B_{2} >0$, all $n_{1}\ge 2$, and all large $z>0$,
\begin{equation}
    \begin{split}\label{power.2.new}
        \mathbb{P}\left(|\textnormal{F}_{j}| \le -\Phi^{-1}(\frac{z^{-2}}{2})\right) \le C \frac{\sqrt{\log{z}}}{\sqrt{n_{1}}|\kappa_{1}|}.
    \end{split}
\end{equation}

By \eqref{power.7.new}, \eqref{power.2.new}, and similar arguments for them, we can deduce that  for all large $n$,
$$ \mathbb{E}(\textnormal{FACT}_{j})\ge \frac{1}{2K}\times \sup_{z\ge 0} \left\{(z - 1) \times \left(1- C  \frac{\sqrt{\log{z}}}{\sqrt{n_{1}} (|\kappa_{1}|\vee |\kappa_{2}|)} \right) \right\},$$
where $C$ has been given in \eqref{power.2.new}. By this result, we can set $z= \exp{(0.25\times C^{-2} \times n_{1}\times (|\kappa_{1}|\vee |\kappa_{2}|)^2)}$ and conclude that for all large $n$,
\begin{equation*}
    \begin{split}
        \mathbb{E}(\textnormal{FACT}_{j}) & \ge K^{-1}\times 0.25\times \exp{(0.25\times C^{-2} \times n_{1}\times (|\kappa_{1}|\vee |\kappa_{2}|)^2)} - 1 \\
        & \ge K^{-1}\times 0.2\times \exp{(0.25\times C^{-2} \times n_{1}\times (|\kappa_{1}|\vee |\kappa_{2}|)^2)}.
    \end{split}    
\end{equation*}
By this result and the assumption that $K$ is a constant, it holds that 
\begin{equation}
    \begin{split}
        \log{\big[\mathbb{E}(\textnormal{FACT}_{j})\big]}
                  & = O\big[ n_{1} \times (|\kappa_{1}|\vee |\kappa_{2}|)^2 - \log{K}\big]\\
                  & = O\left[ \frac{n}{K} \times (|\kappa_{1}|\vee |\kappa_{2}|)^2 \right],
    \end{split}
\end{equation}
 	 which concludes the first assertion of Theorem~\ref{theorem7}. The inclusion of $K$ here is meant to highlight its influence, even though we assume that $K$ is a constant in this context.

We next proceed to establish the second assertion of Theorem~\ref{theorem7}. Using similar arguments as for the second inequality in \eqref{power.7.new}, we can show that for all large $n_{1}$ and all large $z\ge 1$,  
\begin{equation*}
 \begin{split}
     \label{power.7.new.b}
     \textnormal{FACT}_{j} & \ge \frac{1}{2K} \left(\frac{1}{\sqrt{2\Phi(-|\textnormal{F}_{j}|)} } - 1\right) \\     
     & \ge \frac{1}{2K} (z-1)
     \end{split}
 \end{equation*}
with probability at least
\begin{equation*}
\begin{split}
1 - \mathbb{P}\left(\frac{1}{\sqrt{2\Phi(-|\textnormal{F}_{j}|)} } \le z\right)
     & = 1 - \mathbb{P}\left(|\textnormal{F}_{j}| \le -\Phi^{-1}(\frac{z^{-2}}{2})\right)\\
     & \ge 1 - C \frac{\sqrt{\log{z}}}{\sqrt{n_{1}} (|\kappa_{1}| \vee |\kappa_{2}|) },
     \end{split}
\end{equation*}
where the inequality above with constant $C$ follows from \eqref{power.2.new}. This yields the second assertion of Theorem~\ref{theorem7}, which concludes the proof of Theorem~\ref{theorem7}.

 
	
	\renewcommand{\thesubsection}{B.\arabic{subsection}}
	
	\section{Proofs of Propositions \ref{prop1}--\ref{prop5} and some key lemmas} \label{Sec.newB}
	
	\subsection{Proof of Proposition~\ref{prop1}} \label{Sec.newB.1}
	
 To show that $\kappa_{1}$ is lower bounded in the specific setting with \eqref{case.1}, let us write	
	\begin{equation*}
		\kappa_{1} = \mathbb{E}\Big(\mathbb{E}\Big( (Y - \mathbb{E}( Y \vert  \boldsymbol{X}_{-j})) (X_{j} - \mathbb{E} ( X_{j} | \boldsymbol{X}_{-j} )) | \boldsymbol{X}_{-j}\Big)\Big),
	\end{equation*}
	where by the assumptions, these expectations and conditional expectations are well-defined, and the equality above is due to the law of the total expectation. For the inner conditional expectation of $\kappa_{1}$, we can deduce that 
	\begin{equation}\label{prop1.1}
		\begin{split}
			& \mathbb{E}\Big( (Y - \mathbb{E}( Y \vert  \boldsymbol{X}_{-j})) (X_{j} - \mathbb{E} ( X_{j} | \boldsymbol{X}_{-j} )) \big| \boldsymbol{X}_{-j}\Big) \\			
			& = \mathbb{E}[(h(X_{j}) - \mathbb{E} (h(X_{j}) | \boldsymbol{X}_{-j})) (X_{j} - \mathbb{E}(X_{j} | \boldsymbol{X}_{-j}))\ | \boldsymbol{X}_{-j}] \\
			& = \mathbb{E}\Big[\Big(h(X_{j}) - h(\mathbb{E} (X_{j} | \boldsymbol{X}_{-j}))\Big) \Big(X_{j} - \mathbb{E}(X_{j} | \boldsymbol{X}_{-j})\Big)\ \Big| \boldsymbol{X}_{-j}\Big],
	\end{split}\end{equation}
	where the first equality above is due to Condition~\ref{model1} and the last one is because $h(\mathbb{E} (X_{j} | \boldsymbol{X}_{-j}))$ is $\sigma(\boldsymbol{X}_{-j})$-measurable. Our proof relies on the techniques in 
	\cite{cacoullos1989characterizations}, where a lower bound of covariance between $h(X_{j})$ and $X_{j}$ is obtained in terms of the first-order derivative of $h$. Since our technical analysis further takes the conditional expectation into account, we provide a self-contained proof here for completeness.

	In light of  Condition~\ref{model1}, let us define the conditional density of $X_{j}$ given $\boldsymbol{X}_{-j} = \boldsymbol{z}$ as $f_{X_{j}| \boldsymbol{z}}(x)$ such that 
	\[ f_{X_{j}| \boldsymbol{z}}(x) \times  f_{\boldsymbol{X}_{-j}}(\boldsymbol{z})  = f_{X_{j}}(x), \]
	where $f_{X_{j}}$ and $f_{\boldsymbol{X}_{-j}}$ denote the density functions of the distributions of $X_{j}$ and $\boldsymbol{X}_{-j}$, respectively. In addition, we denote the versions of $\mathbb{E} (X_{j} | \boldsymbol{X}_{-j})$ and the RHS of (\ref{prop1.1}) as $\mu( \boldsymbol{X}_{-j})$ and $\nu(\boldsymbol{X}_{-j})$, respectively, for some measurable functions $\mu$ and $\nu$. We will derive an expression for the RHS of (\ref{prop1.1}) in terms of $\partial h(x) / \partial x$.  By the change of variable formula, for each real vector $\boldsymbol{z}\in [0, 1]^{p-1}$ we have 
	\[\nu(\boldsymbol{z}) = \int_{-\infty}^{\infty} (h(x) - h(\mu(\boldsymbol{z}) ))  (x - \mu(\boldsymbol{z}) ) f_{X_{j}| \boldsymbol{z}}(x) dx.\]	
	Thus, it holds almost surely that 
	\begin{equation}\label{prop1.2}
		\begin{split}
			\textnormal{RHS of (\ref{prop1.1})}& = \nu(\boldsymbol{X}_{-j}) \\
			& = \int_{-\infty}^{\infty} (h(x) - h(\mu(\boldsymbol{X}_{-j}) )) (x - \mu(\boldsymbol{X}_{-j}) ) f_{X_{j}| \boldsymbol{X}_{-j}}(x) dx.
	\end{split}\end{equation}
	
	Recall the assumption that the derivative of $h$ is integrable and denote by $\boldsymbol{1}$ the indicator function. Then we can resort to the fundamental theorem of calculus~\citep{botsko1986stronger} to deduce that for $x\in\mathbb{R}$,
	\begin{align*}
	h(x) - h(\mu(\boldsymbol{X}_{-j}))& = \left(\int_{\mu(\boldsymbol{X}_{-j})}^{x}  \frac{\partial h(t)}{\partial t}dt \right) \boldsymbol{1}_{x \ge \mu(\boldsymbol{X}_{-j})} \\
	&\quad+ \left(\int_{x}^{\mu(\boldsymbol{X}_{-j})}  \frac{\partial h(t)}{\partial t} dt \right) \boldsymbol{1}_{x \le  \mu(\boldsymbol{X}_{-j})},
	\end{align*}
	which is equal to zero at $x = \mu(\boldsymbol{X}_{-j})$. It follows from such a representation that 
	\begin{equation}\label{prop1.3}
		\begin{split}
			& \textnormal{RHS of (\ref{prop1.2}) } \\
			&= \int_{-\infty}^{\infty} \left(\int_{\mu(\boldsymbol{X}_{-j})}^{x}  \frac{\partial h(t)}{\partial t}dt \right) \boldsymbol{1}_{x \ge \mu(\boldsymbol{X}_{-j})} (x - \mu(\boldsymbol{X}_{-j}) )  f_{X_{j}| \boldsymbol{X}_{-j}}(x) dx \\
			& \qquad + \int_{-\infty}^{\infty} \left(\int_{x}^{\mu(\boldsymbol{X}_{-j})}  \frac{\partial h(t)}{\partial t} dt \right) \boldsymbol{1}_{x\le \mu(\boldsymbol{X}_{-j})} (\mu(\boldsymbol{X}_{-j}) - x)   f_{X_{j}| \boldsymbol{X}_{-j}}(x) dx \\
			& = \int_{-\infty}^{\infty} \left(\int_{-\infty}^{\infty} \boldsymbol{1}_{t \in [\mu(\boldsymbol{X}_{-j}), x]}  \frac{\partial h(t)}{\partial t} dt \right)  (x - \mu(\boldsymbol{X}_{-j}) ) f_{X_{j}| \boldsymbol{X}_{-j}}(x) dx \\
			& \qquad+ \int_{-\infty}^{\infty} \left(\int_{-\infty}^{\infty} \boldsymbol{1}_{t \in [x, \mu(\boldsymbol{X}_{-j})]}  \frac{\partial h(t)}{\partial t} dt \right)  (\mu(\boldsymbol{X}_{-j}) - x)  f_{X_{j}| \boldsymbol{X}_{-j}}(x) dx.
	\end{split}\end{equation}
	
	Since the derivative is bounded in absolute value and $0\le X_{j} \le 1$, we can show that the two integrations on the RHS of (\ref{prop1.3}) are absolutely integrable. Hence, an application of  Fubini's theorem and the facts that $\boldsymbol{1}_{t \in [\mu(\boldsymbol{X}_{-j}), x]}  = \boldsymbol{1}_{x\ge t}\boldsymbol{1}_{t \ge \mu(\boldsymbol{X}_{-j})} $ and $\boldsymbol{1}_{t \in [x, \mu(\boldsymbol{X}_{-j})]}  = \boldsymbol{1}_{x\le  t}\boldsymbol{1}_{t \le \mu(\boldsymbol{X}_{-j})} $ yields that  
	\begin{equation}
		\begin{split}\label{prop1.4}
			& \textnormal{RHS of (\ref{prop1.3}) } \\
		&=\int_{-\infty}^{\infty}  \int_{-\infty}^{\infty} \boldsymbol{1}_{t \in [\mu(\boldsymbol{X}_{-j}), x]}   \left( \frac{\partial h(t)}{\partial t} \right) (x - \mu(\boldsymbol{X}_{-j}) )  f_{X_{j}| \boldsymbol{X}_{-j}}(x) dx dt  \\
			& \qquad+ \int_{-\infty}^{\infty}  \int_{-\infty}^{\infty} \boldsymbol{1}_{t \in [x, \mu(\boldsymbol{X}_{-j})]} \left( \frac{\partial h(t)}{\partial t} \right) (\mu(\boldsymbol{X}_{-j}) - x)  f_{X_{j}| \boldsymbol{X}_{-j}}(x) dx dt\\
			& = \int_{-\infty}^{\infty}  \int_{t}^{\infty}  (x - \mu(\boldsymbol{X}_{-j}) )  f_{X_{j}| \boldsymbol{X}_{-j}}(x) dx \left( \frac{\partial h(t)}{\partial t} \right) \boldsymbol{1}_{t \ge \mu(\boldsymbol{X}_{-j})} dt \\
			& \qquad + \int_{-\infty}^{\infty}  \int_{-\infty}^{t}  (\mu(\boldsymbol{X}_{-j}) - x)  f_{X_{j}| \boldsymbol{X}_{-j}}(x) dx \left( \frac{\partial h(t)}{\partial t} \right) \boldsymbol{1}_{t \le \mu(\boldsymbol{X}_{-j})} dt.
		\end{split}
	\end{equation}
	
	Furthermore, since 
	\[\int_{t}^{\infty}  (x - \mu(\boldsymbol{X}_{-j}) )  f_{X_{j}| \boldsymbol{X}_{-j}}(x) dx = \int_{-\infty}^{t}  (\mu(\boldsymbol{X}_{-j}) - x)  f_{X_{j}| \boldsymbol{X}_{-j}}(x) dx,\]
	we see that the RHS of (\ref{prop1.4}) becomes
	\begin{equation}
		\begin{split}\label{prop1.5}
			& \textnormal{RHS of (\ref{prop1.4}) } = \int_{-\infty}^{\infty}   \frac{\partial h(t)}{\partial t}  \left(\int_{-\infty}^{t}  (\mu(\boldsymbol{X}_{-j}) - x)  f_{X_{j}| \boldsymbol{X}_{-j}}(x) dx   \right) dt.
	\end{split}\end{equation} 
	Let us set $h(x) = x$ in (\ref{prop1.5}). Then from \eqref{prop1.1}, we have 
	\begin{equation}\label{prop1.6}
		\mathbb{E}(\textnormal{Var}(X_{j}| \boldsymbol{X}_{-j})) = \mathbb{E} \left[\int_{-\infty}^{\infty}    \left(\int_{-\infty}^{t}  (\mu(\boldsymbol{X}_{-j}) - x)  f_{X_{j}| \boldsymbol{X}_{-j}}(x) dx   \right) dt \right].
	\end{equation}
	Therefore, it follows from the assumption that $h$ is monotonic and (\ref{prop1.1})--(\ref{prop1.6}) that 
	\begin{equation*}
		\begin{split}
	|\kappa_{1}| & = \mathbb{E}\left[\int_{-\infty}^{\infty}   \left|\frac{\partial h(t)}{\partial t}\right|  \left(\int_{-\infty}^{t}  (\mu(\boldsymbol{X}_{-j}) - x)  f_{X_{j}| \boldsymbol{X}_{-j}}(x) dx   \right) dt \right] \\
	&\ge \left(\inf_{x\in [0, 1]} \left|\frac{\partial h(x)}{\partial x}\right| \right)\mathbb{E}(\textnormal{Var}(X_{j}| \boldsymbol{X}_{-j})),
	\end{split}
	\end{equation*}
	which completes the proof of Proposition~\ref{prop1}.
	
	\subsection{Proof of Proposition~\ref{prop2}} \label{Sec.newB.2}
	
 From Condition~\ref{model1}, $g_{1l}(x)$'s and $g_{2l}(x)$'s given as in \eqref{case.1}, and the distributional assumption of $\boldsymbol{X}$, it holds that 
	\begin{equation}
		\label{prop2.1}
		\begin{split}
		\kappa_{l}  & = \mathbb{E}\big((Y - \mathbb{E}( Y \vert  \boldsymbol{X}_{-j})) (X_{j}^{l} - \mathbb{E} X_{j}^{l} )\big) \\
		&= \mathbb{E}( h(X_{j})  (X_{j}^{l} - \mathbb{E} X_{j}^{l}) ),
		\end{split}
	\end{equation}
	where the conditional expectation $\mathbb{E}(H(\boldsymbol{X}_{-j}) + \varepsilon| \boldsymbol{X}_{-j})$ is well-defined since we assume that the first moments of $H(\boldsymbol{X}_{-j})$ and $\varepsilon$ exist. By (\ref{prop2.1}), the distributional assumption of $\boldsymbol{X}$, and the form of $h$, we can deduce that for $1\le l\le L$,
	\begin{equation}
		\begin{split}\label{prop2.2}
			\kappa_{l} & = (a_{1}, \dots, a_{L})  (\mathbb{E} X_{j}^{1+l} - (\mathbb{E} X_{j}) (\mathbb{E} X_{j}^{l}),  \dots, \mathbb{E} X_{j}^{L+l} - (\mathbb{E} X_{j}^{L}) (\mathbb{E} X_{j}^{l}) )^{\top}\\
			& = (a_{1}, \dots, a_{L})\\
			&\hspace{2em}\times \left(\frac{1}{l+2} - \left(\frac{1}{2}\right) \left(\frac{1}{l+1} \right), \dots, \frac{1}{L + l + 1} - \left( \frac{1}{L + 1}\right)\left(\frac{1}{l+1} \right)\right)^{\top},
		\end{split}
	\end{equation}
	where we have used the fact that $\mathbb{E} X_{j}^{k} = (k+1)^{-1}$ due to the distributional assumption.
	
	Let us define an $L\times L$ matrix \[ D \coloneqq [(i+j+1)^{-1}]_{i,j=1, \dots, L} \] 
	and $L$-dimensional vectors $B = C \coloneqq (\frac{1}{2}, \dots, \frac{1}{L+1})^{\top}$. Then by (\ref{prop2.2}), we have 
	\begin{equation}\label{prop2.3}
		(\kappa_{1}, \dots, \kappa_{L}) = (a_{1}, \dots, a_{L}) (D - CB^{\top}).
	\end{equation}
We will show that $D - CB^{\top}$ is positive definite. To this end, let us introduce a Hilbert matrix of order $L + 1$
	\[
	[(i+j-1)^{-1}]_{i,j = 1,\dots, L+1} = \left(\begin{array}{cc}
		1 & B^{\top}\\
		C & D
	\end{array}\right).\]
	Since a Hilbert matrix is positive definite~\citep{choi1983tricks}, its inverse exists and is also positive definite. With the aid of the block matrix inversion formula, we see that $(D - CB^{\top})^{-1}$ is the bottom-right block of the inverse of the Hilbert matrix and hence is positive definite. This entails that $D - CB^{\top}$ is positive definite. Thus, it follows from this result and (\ref{prop2.3}) that if $\sum_{l}|a_{l}|>0$, we have 
	\begin{equation*}
		\sum_{l=1}^{L}|\kappa_{l}| \ge \left(\sum_{l=1}^{L}|a_{l}|\right)^{-1} \left(\sum_{l=1}^{L}\kappa_{l}a_{l}\right) \ge  \frac{\sum_{l=1}^{L}|a_{l}|}{L}\lambda_{\min}\left( D - CB^{\top}\right)> 0,
	\end{equation*}
	where in the second inequality above, we have used the fact that  $\sum_{l=1}^{L}|a_{l}| \le  \sqrt{L\sum_{l=1}^{L}|a_{l}|^2}$. By this result, we can conclude the first assertion of Proposition~\ref{prop2}.
	
	Next, a direct calculation shows that the minimum eigenvalues of $D - CB^{\top}$ is larger than $0.002$ for $L \le  2$, and hence for $L= 2$, we have that 
	$$ |\kappa_{1}| + |\kappa_{2}|  \ge 0.001\times (|a_{1}| + |a_{2}|),$$
	which concludes the proof of Proposition~\ref{prop2}.

\subsection{Proof of Proposition~\ref{prop5}}\label{Sec.B.3}

For the first assertion, we can deduce that
\begin{equation}
    \begin{split}
        \mathbb{E} (\frac{\#(\widetilde{S}\cap \mathcal{H}_{0})}{\# \widetilde{S} \vee 1}) & = \mathbb{E} (\frac{ \sum_{j\in \mathcal{H}_{0}} \boldsymbol{1}_{j \in \widetilde{S} } }{\widetilde{k} \vee 1}) \\
        & \le \mathbb{E} ( \sum_{j\in \mathcal{H}_{0}} \boldsymbol{1}_{j \in \widetilde{S} } \times \widetilde{\textnormal{FACT}}_{j}\times \alpha \times p^{-1} ) \\
        & \le \alpha \times p^{-1}\times \mathbb{E} ( \sum_{j\in \mathcal{H}_{0}}  \widetilde{\textnormal{FACT}}_{j} ) \\
        & \le \alpha \times p^{-1}\times \mathbb{E} ( \sum_{j\in \mathcal{H}_{0}}  \textnormal{FACT}_{j} )\\
        & \le \alpha,
    \end{split}
\end{equation}
where the first inequality above holds because  $\widetilde{\textnormal{FACT}}_{j}\times \alpha \times p^{-1} \ge \widetilde{k}^{-1}$ for each $j\in \widetilde{S}$, and the third inequality above follows from the definition of $\widetilde{\textnormal{FACT}}_{j}$'s in \eqref{new.fact.1} and the fact that $\textnormal{FACT}_{j}$, $\textnormal{FACT}_{j, 1}$, $\dots, \textnormal{FACT}_{j, B}$ have the same distribution. This finishes the proof for the first assertion of Proposition~\ref{prop5}.

Next, in view of the definition of \eqref{new.fact.1}, we have $k\in \widehat{S}$ if $k\in \widetilde{S} = \{j: \widetilde{\textnormal{FACT}}_{j} \ge \widetilde{\textnormal{FACT}}_{(\widetilde{k})}\}$. By this result and definition \eqref{new.fact.1}, it holds that $\widetilde{E}_{k} \ge \widetilde{\textnormal{FACT}}_{k}$ for each $k \in \widetilde{S}$. Therefore, it follows that 
$$\widetilde{k} \times \min_{k \in \widetilde{S}} \widetilde{E}_{k} \ge \widetilde{k} \times \widetilde{\textnormal{FACT}}_{(\widetilde{k})} \ge p/ \alpha,$$ 
which entails that $\widetilde{S} \subset S^{\dagger}$ according to the definition of $S^{\dagger}$ in \eqref{reproducible.fact.1}. This completes the proof of Proposition~\ref{prop5}.

	\subsection{Lemma~\ref{claim2} and its proof} \label{Sec.newC.2}
	
	
	All the notation here is the same as in the proof of Theorem~\ref{theorem3}. In particular, recall that $A_{1i} = (g_{1}(Y_{i}) - \mathbb{E}(g_{1}(Y_{i}) | \boldsymbol{X}_{-ij} ) ) (g_{2}(X_{ij}) - \mathbb{E} (g_{2}(X_{j})| \boldsymbol{X}_{-ij}) )$ and $\sigma_{j}^{2} = \textnormal{Var}(A_{11})$.
	\begin{lemma}\label{claim2}
		Let $X_{j}$ be a null feature and assume that $\mathbb{E} [g_{1}(Y)]^{4} < \infty$, $0 <\sigma_{j}^{2} < \infty$, and $0 \le g(X_{j})\le 1$. Then there exists some $C>0$ such that for each positive integer $n$,
		\[\sup_{t\in\mathbb{R}} |\mathbb{P}(n^{-1/2}\sum_{i=1}^{n} \frac{A_{1i}}{\sigma_{j}} \le t) - \Phi(t)|\le C(16\mathbb{E} [g_{1}(Y)]^{4})^{1/3}\sigma_{j}^{-4/3}n^{-1/3}, \]
		where $\Phi(\cdot)$ stands for the cumulative distribution function of the standard Gaussian distribution. Note that constant $C$ does not depend on $\mathbb{E} [g_{1}(Y)]^{4}$, $\sigma_{j}^{2}$, or index $1\le j\le p$.
	\end{lemma}
	
	\noindent \textit{Proof}. The proof of this lemma involves an application of the central limit theorem~\citep{le1986central, petrov1977sums}. To do so, we will verify the required conditions. Since $X_{j}$ is a null feature and the observations are i.i.d., we have that  $\mathbb{E} A_{1i} = 0$ and $\sigma_{j}^{2} = \mathbb{E} (A_{11}^2)$ for each $i$. Let $s \coloneqq \sqrt{n}\sigma_{j}$ and $\varepsilon = 16^{1/3}(\mathbb{E} [g_{1}(Y)]^{4})^{1/3} \sigma_{j}^{-4/3}n^{-1/3}$. Then we can deduce that 
	\begin{equation*}\begin{split}	
			\label{clt1}
			\sum_{i=1}^{n}\mathbb{E}\left[ (\frac{A_{1i}}{s})^{2}\boldsymbol{1}_{ |A_{1i}| > s\varepsilon}\right] &  = \frac{1}{n\sigma_{j}^{2} } \sum_{i=1}^{n} \mathbb{E} \left( A_{1i}^{2} \boldsymbol{1}_{|A_{1i}| > s\varepsilon} \right) \\&= \sigma_{j}^{-2} \mathbb{E} (A_{11}^{2} \boldsymbol{1}_{|A_{11}| > s\varepsilon})\\
			& \le \sigma_{j}^{-2} (\mathbb{E} A_{11}^{4})^{1/2} \{\mathbb{P}(|A_{11}| > s \varepsilon)\}^{1/2}  \\
			&\le \sigma_{j}^{-2} (\mathbb{E} A_{11}^{4})^{1/2} \left(\frac{\mathbb{E} A_{11}^{4}} {|s\varepsilon|^{4}}\right)^{1/2} \\	
			& \le \varepsilon,
		\end{split}
	\end{equation*}
	where $\boldsymbol{1}$ represents the indicator function, the second equality above is due to the assumption of i.i.d. observations, the first inequality above is due to the Cauchy--Schwartz inequality, the second one is an application of the Markov inequality, and the last one is entailed by the assumption that $0 \le g_{2}(X_{j})\le 1$ as well as some simple calculations. Therefore, we can resort to the central limit theorem~\citep{le1986central, petrov1977sums} to obtain that for some $C>0$,
	\[\sup_{t\in\mathbb{R}} |\mathbb{P}( n^{-1/2}\sum_{i=1}^{n} \frac{A_{1i}}{\sigma_{j}} \le t) - \Phi(t)| \le C \varepsilon.\]
	This concludes the proof of  Lemma~\ref{claim2}.
	
\subsection{Lemma~\ref{lower.var.bounds} and its proof} \label{Sec.newC.32}
All the notation here is the same as in the main body of the paper.
	\begin{lemma}
	\label{lower.var.bounds}
Assume that $\textnormal{Var}(g_{2}(X_{j}) | \boldsymbol{X}_{-j}) \ge \varsigma_{1}$ and $\textnormal{Var}(g_{1}(Y)|\boldsymbol{X})\ge \varsigma_{2}$ almost surely for some $0\le g_{2}(X_{j})\le 1$ and $\varsigma_{1}, \varsigma_{2}>0$. Then we have 
	\begin{equation*}
	    \begin{split}
	        \textnormal{Var}\{[ g_{1}(Y) - \mathbb{E}(g_{1}(Y)|\boldsymbol{X}_{-j}) ]  [g_{2}(X_{j}) - \mathbb{E}(g_{2}(X_{j}) |\boldsymbol{X}_{-j} )  ]\} & \ge \frac{1}{16}\varsigma_{2}\varsigma_{1}^2.
	    \end{split}
	\end{equation*}
	\end{lemma}
	
	\noindent\textit{Proof}
	For any $G(\boldsymbol{X})$ with $\mathbb{P}(|G(\boldsymbol{X})|> d)> \delta$ for some $d, \delta>0$, it holds that \begin{equation}
	    \begin{split}\label{lower.var.bounds.1}
	        &\textnormal{Var}\big\{(g_{1}(Y) - \mathbb{E}(g_{1}(Y)|\boldsymbol{X}_{-j})) G(\boldsymbol{X}) \big\}  \\
	        &\ge \mathbb{E}\big\{\textnormal{Var}\big[(g_{1}(Y) - \mathbb{E}(g_{1}(Y)|\boldsymbol{X}_{-j})) G(\boldsymbol{X}) |\boldsymbol{X}\big] \big\} \\
	        & = \mathbb{E}\big\{\textnormal{Var}\big[(g_{1}(Y)  - \mathbb{E}(g_{1}(Y)|\boldsymbol{X}) + \mathbb{E}(g_{1}(Y)|\boldsymbol{X}) - \mathbb{E}(g_{1}(Y)|\boldsymbol{X}_{-j}) ) G(\boldsymbol{X}) | \boldsymbol{X}\big] \big\}\\
	        & = \mathbb{E}\Big\{ \Big[\big[g_{1}(Y)  - \mathbb{E}(g_{1}(Y)|\boldsymbol{X}) + \mathbb{E}(g_{1}(Y)|\boldsymbol{X})  - \mathbb{E}(g_{1}(Y)|\boldsymbol{X}_{-j}) \big] G(\boldsymbol{X}) \\
        & \qquad - \mathbb{E}\Big[ \big[g_{1}(Y) - \mathbb{E}(g_{1}(Y)|\boldsymbol{X}) + \mathbb{E}(g_{1}(Y)|\boldsymbol{X}) - \mathbb{E}(g_{1}(Y)|\boldsymbol{X}_{-j}) \big] G(\boldsymbol{X})\Big| \boldsymbol{X}\Big] \Big]^2\Big\} \\
	        & = \mathbb{E}\big\{ \big[( g_{1}(Y)  - \mathbb{E}(g_{1}(Y)|\boldsymbol{X})) G(\boldsymbol{X}) \big]^2\big\}\\
	        &\ge \mathbb{E}\big\{ \big[( g_{1}(Y) - \mathbb{E}(g_{1}(Y)|\boldsymbol{X})) d \big]^2 \boldsymbol{1}_{|G(\boldsymbol{X})| >d }\big\} \\
	        & \ge \varsigma_{2}d^2\delta,
	    \end{split}
	\end{equation}
	where the third equality above follows from the facts that  
 $$\mathbb{E}\big\{ \big[ \mathbb{E}(g_{1}(Y)|\boldsymbol{X}) - \mathbb{E}(g_{1}(Y)|\boldsymbol{X}_{-j}) \big] G(\boldsymbol{X})| \boldsymbol{X}\big\} =\big[ \mathbb{E}(g_{1}(Y)|\boldsymbol{X}) - \mathbb{E}(g_{1}(Y)|\boldsymbol{X}_{-j}) \big] G(\boldsymbol{X})$$ almost surely and $\mathbb{E}\big\{ \big[g_{1}(Y) - \mathbb{E}(g_{1}(Y)|\boldsymbol{X})\big] G(\boldsymbol{X})|\boldsymbol{X}\big\} = 0$, and the last inequality above is due to the assumption of  $\textnormal{Var}(g_{1}(Y)|\boldsymbol{X})\ge \varsigma_{2}$.

	Next, for each Borel set $\mathcal{A} \in \mathcal{R}$ and  $G(\boldsymbol{X})$ such that $-1 \le G(\boldsymbol{X})\le 1$ and $\mathbb{E}(G(\boldsymbol{X})) = 0$, we have that 
	\begin{equation*}
	    \begin{split}
	        \textnormal{Var}(G(\boldsymbol{X}))&= \mathbb{E}\big\{ \big[G(\boldsymbol{X}) - \mathbb{E}(G(\boldsymbol{X})) \big]^2 \boldsymbol{1}_{G(\boldsymbol{X})\in \mathcal{A} } + \big[G(\boldsymbol{X}) - \mathbb{E}(G(\boldsymbol{X})) \big]^2 \boldsymbol{1}_{G(\boldsymbol{X})\in \mathcal{A}^c }\big\}\\
	        & \le (\sup \mathcal{A} - \inf \mathcal{A})^2 \mathbb{P}(G(\boldsymbol{X})\in \mathcal{A}) + \mathbb{P}(G(\boldsymbol{X})\in \mathcal{A}^c),
	    \end{split}
	\end{equation*}
	and thus 
	$$\mathbb{P}(G(\boldsymbol{X})\in \mathcal{A})\le \frac{ 1 - \textnormal{Var}(G(\boldsymbol{X}))}{1 - (\sup \mathcal{A} - \inf \mathcal{A})^2}.$$
	This result in combination with setting $\mathcal{A} = [-\sqrt{\frac{\varsigma_{1}}{8}}, \sqrt{\frac{\varsigma_{1}}{8}}]$ and an additional assumption that $\textnormal{Var}(G(\boldsymbol{X}))\ge \varsigma_{1}$ leads to
	\begin{equation}
	    \label{lower.var.bounds.2}
	\mathbb{P}(|G(\boldsymbol{X})| > \sqrt{\frac{\varsigma_{1}}{8}})>  \frac{1}{2} \varsigma_{1}.
	\end{equation}
	Therefore, from \eqref{lower.var.bounds.1}--\eqref{lower.var.bounds.2}, the assumption of  $\textnormal{Var}(g_{2}(X_{j}) | \boldsymbol{X_{-j}})\ge \varsigma_{1}$, and letting  $G(\boldsymbol{X}) = g_{2}(X_{j}) - \mathbb{E}(g_{2}(X_{j})| \boldsymbol{X}_{-j})$, we can obtain the desired conclusion of Lemma~\ref{lower.var.bounds}. This completes the proof of Lemma~\ref{lower.var.bounds}.
	
 

\end{document}